%% file: main.tex
\documentclass[letterpaper]{article} % DO NOT CHANGE THIS
\usepackage{aaai2026}  % DO NOT CHANGE THIS #submission/camera-ready
\usepackage{amsmath}
\usepackage{svg}
\input{preamble}

\usepackage{xcolor}
\definecolor{mycol}{RGB}{251,49,153}

\usepackage{subfig}
\title{\texttt{O3SLM}: Open Weight, Open Data, and Open Vocabulary Sketch-Language Model}
\author{
    Rishi Gupta\textsuperscript{\rm 1},
    Mukilan Karuppasamy\textsuperscript{\rm 1}\equalcontrib,
    Shyam Marjit\textsuperscript{\rm 1}\equalcontrib,
    Aditay Tripathi\textsuperscript{\rm 1}\footnote{Major contributions made during association with IISc.},\\
    Anirban Chakraborty\textsuperscript{\rm 1}
}
\affiliations {
    \textsuperscript{\rm 1}Indian Institute of Science, Bangalore\\
    rishig@iisc.ac.in, mukilan.nitt@gmail.com, shyammarjit@iisc.ac.in, aditaytr@gmail.com, anirban@iisc.ac.in
}

\usepackage{bibentry}
\begin{document}

% \maketitle
% \twocolumn[{%
% \renewcommand\twocolumn[1][]{#1}%
\maketitle

\begin{center}

\end{center}
\input{sec/0_abstact}

% Uncomment the following to link to your code, datasets, an extended version or similar.
% You must keep this block between (not within) the abstract and the main body of the paper.
% Project Page: {\color{mycol}\url{}}
\begin{links}
    % \link{Code}{https://aaai.org/example/code}
    % \link{Datasets}{https://aaai.org/example/datasets}
    % \link{Extended version}{https://arxiv.org/abs/2511.14368}
    \link{Project page}{https://vcl-iisc.github.io/O3SLM/}
\end{links}

\input{sec/1_introduction}
\input{sec/2_releated_work}

\input{sec/3_dataset}

\input{sec/4_methods}
\input{sec/5_experiments}
\input{sec/6_conclusion}
\input{sec/acknowledgements}

% \section{Acknowledgments}

% \cleardoublepage

\bibliography{aaai2026}

\begin{center}
    {\Large \textbf{Supplementary}}
\end{center}

\input{sec/append}

% % \input{ReproducibilityChecklist/LaTeX/ReproducibilityChecklist}
\end{document}

%% file: preamble.tex
\usepackage{subcaption}
\usepackage{times}  % DO NOT CHANGE THIS
\usepackage{helvet}  % DO NOT CHANGE THIS
\usepackage{courier}  % DO NOT CHANGE THIS
\usepackage[hyphens]{url}  % DO NOT CHANGE THIS
\usepackage{graphicx} % DO NOT CHANGE THIS
\usepackage{natbib}
\usepackage{tabularx}% DO NOT CHANGE THIS AND DO NOT ADD ANY OPTIONS TO IT
\usepackage{caption} % DO NOT CHANGE THIS AND DO NOT ADD ANY OPTIONS TO IT
\usepackage{algorithm}
\usepackage{algorithmic}
\usepackage{booktabs} 
\usepackage{multirow}

\usepackage{newfloat}
\usepackage{listings}

\DeclareCaptionStyle{ruled}{labelfont=normalfont,labelsep=colon,strut=off} % DO NOT CHANGE THIS
\floatstyle{ruled}
\newfloat{listing}{tb}{lst}{}
\floatname{listing}{Listing}
\usepackage{siunitx, booktabs}
\usepackage{adjustbox}
\usepackage[table]{xcolor}
\definecolor{molmocolor}{RGB}{240, 82, 156}
\definecolor{tablegray}{RGB}{223, 242, 252}

\newcolumntype{x}[1]{>{\centering\arraybackslash}p{#1pt}}
\newcolumntype{y}[1]{>{\raggedright\arraybackslash}p{#1pt}}
\newcolumntype{z}[1]{>{\raggedleft\arraybackslash}p{#1pt}}

\usepackage{color}
\definecolor{citecolor}{RGB}{34,139,34}
\definecolor{citecolor2}{HTML}{0071bc}
\definecolor{lightred}{RGB}{241,140,142}
\definecolor{carmine}{rgb}{0.59, 0.0, 0.09}

\usepackage{pifont} 
\usepackage{tabularx}
\newcommand{\cmark}{\ding{51}}%
\newcommand{\xmark}{\ding{55}}%

\newlength\savewidth\newcommand\shline{\noalign{\global\savewidth\arrayrulewidth
		\global\arrayrulewidth 1pt}\hline\noalign{\global\arrayrulewidth\savewidth}}

\usepackage{booktabs,arydshln}
\makeatletter

\def\adl@drawiv#1#2#3{%
        \hskip.5\tabcolsep
        \xleaders#3{#2.5\@tempdimb #1{1}#2.5\@tempdimb}%
                #2\z@ plus1fil minus1fil\relax
        \hskip.5\tabcolsep}
\newcommand{\cdashlinelr}[1]{%
  \noalign{\vskip\aboverulesep
           \global\let\@dashdrawstore\adl@draw
           \global\let\adl@draw\adl@drawiv}
  \cdashline{#1}
  \noalign{\global\let\adl@draw\@dashdrawstore
           \vskip\belowrulesep}}

\newcommand{\cboldline}[1]{%
  \noalign{\global\setlength{\saved@arrayrulewidth}{\arrayrulewidth}%
           \global\setlength{\arrayrulewidth}{0.7pt}}% ← Thicker line here
  \cline{#1}
  \noalign{\global\setlength{\arrayrulewidth}{\saved@arrayrulewidth}}}

\makeatother

\usepackage{makecell}

\makeatletter

%% file: sec/0_abstact.tex
\begin{abstract}

While Large Vision Language Models (LVLMs) are increasingly deployed in real-world applications, their ability to interpret abstract visual inputs remains limited. Specifically, they struggle to comprehend hand-drawn sketches, a modality that offers an intuitive means of expressing concepts that are difficult to describe textually. We identify the primary bottleneck as the absence of a large-scale dataset that jointly models sketches, photorealistic images, and corresponding natural language instructions. To address this, we present two key contributions: (1) a new, large-scale dataset of image-sketch-instruction triplets designed to facilitate both pretraining and instruction tuning, and (2) \textbf{\texttt{O3SLM}}, an LVLM trained on this dataset. Comprehensive evaluations on multiple sketch-based tasks: (a) object localization, (b) counting, (c) image retrieval \emph{i.e.}, (SBIR and fine-grained SBIR), and (d) visual question answering (VQA); while incorporating the three existing sketch datasets, namely QuickDraw!, Sketchy, and Tu Berlin, along with our generated \textbf{\texttt{SketchVCL}} dataset, show that \textbf{\texttt{O3SLM}} achieves state-of-the-art performance, substantially outperforming existing LVLMs in sketch comprehension and reasoning.

\end{abstract}

%% file: sec/1_introduction.tex
\section{1. Introduction}
\label{sec:intro}

\begin{figure}[!ht]
    \centering
    \includegraphics[width=\linewidth]{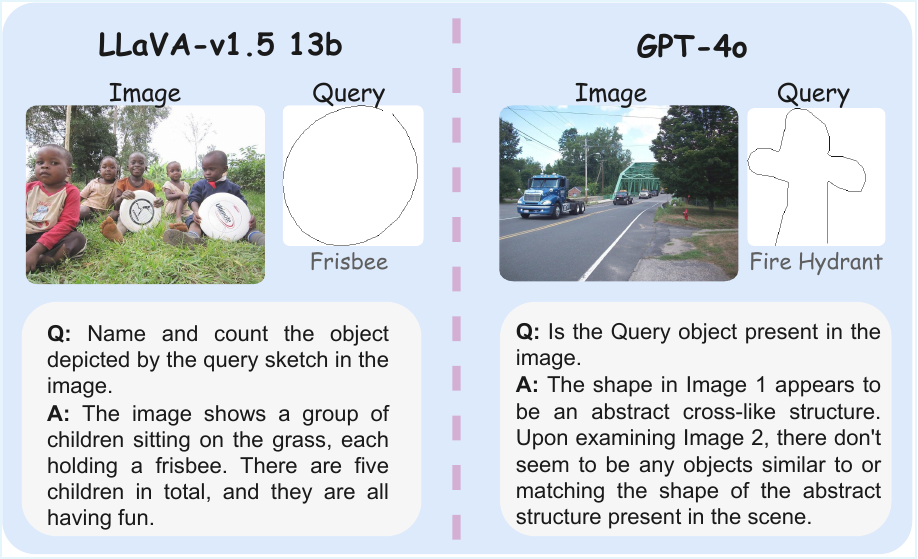}
    \caption{\textbf{Limitations of LVLMs in Sketch Understanding.} Although current LVLMs can interpret sketches to some level of abstraction, they struggle in sketch understanding for downstream tasks like detection and reasoning.}
    \label{fig:motivation}
\end{figure}

\begin{figure*}[t]
    \centering
    \includegraphics[width=\linewidth]{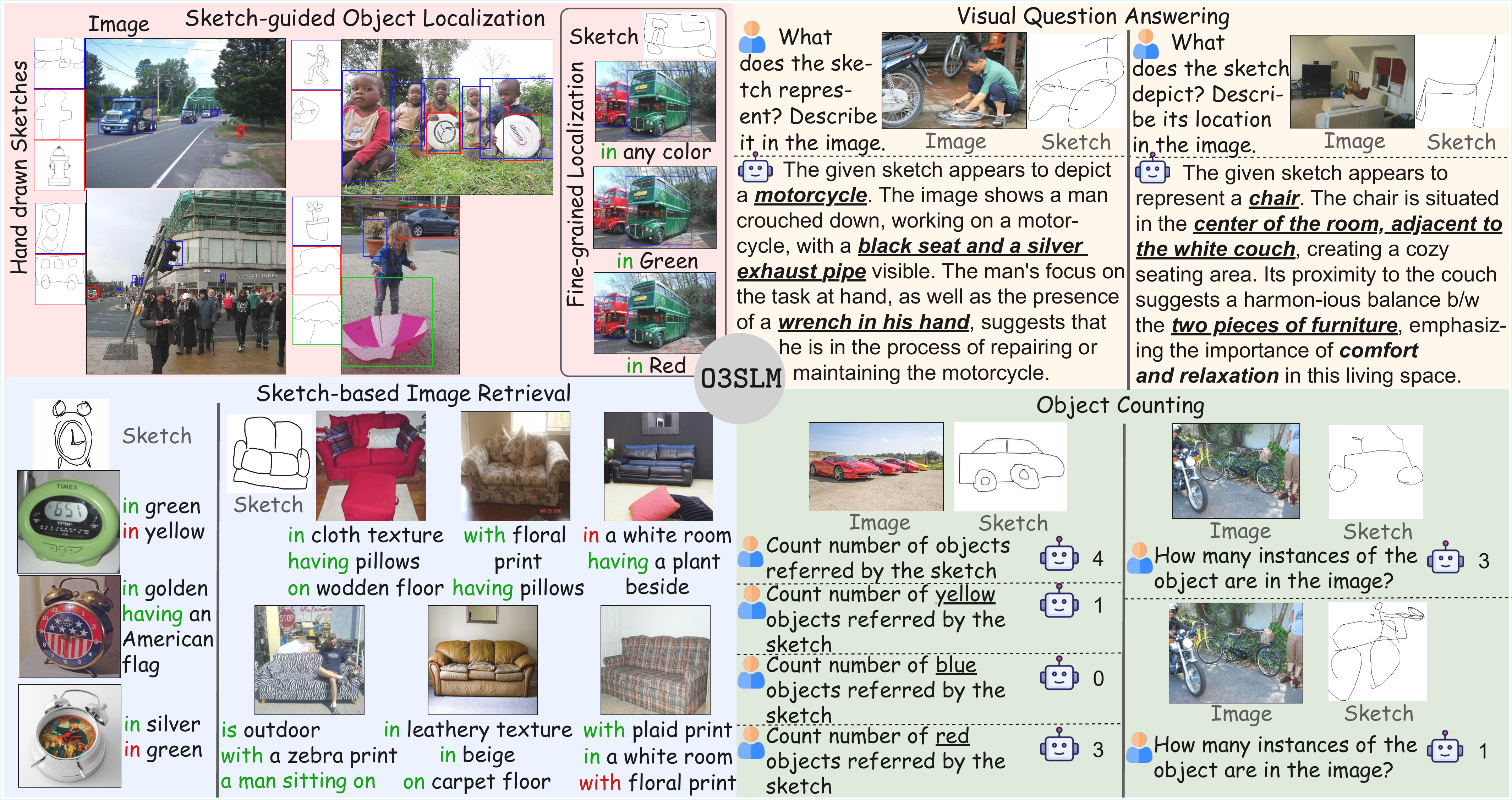}
    \caption{\textbf{Capabilities of our model - \texttt{O3SLM.}} Our model is the first Large Vision-Language Model (LVLM) to demonstrate advanced alignment between sketches, images, and text—where existing LVLMs consistently fail (see Table~\ref{tab:benchmark_results_count}). Through extensive pretraining on our proposed \textbf{\texttt{SketchVCL}} dataset, the model develops a robust understanding of crude hand-drawn sketches and how they relate to the visual and textual modalities in which current LVLMs already excel. This training enables cross-modal transfer, allowing the model to handle fine-grained queries using sketch-text pairs, even though it was originally trained with sketches alone. \textbf{\texttt{O3SLM}} is trained across multiple tasks, including Visual Question Answering (VQA), Sketch-Based Image Retrieval (SBIR), sketch-based counting, and sketch-based object detection.}
    \label{fig:main_fig}
    
\end{figure*}

The landscape of Large Vision Language Models (LVLMs) is advancing rapidly, with models becoming increasingly powerful and accessible. Foundational open-source models such as LLaVA~\cite{liu2023llava}, Qwen-VL~\cite{wang2024qwen2}, DeepSeek-VL2~\cite{wu2024deepseekvl2}, and Molmo~\cite{deitke2025molmo} have demonstrated remarkable vision-language capabilities on benchmark tasks like Visual Question Answering (VQA) and Document VQA. This success has catalyzed a wave of development in fine-grained LVLMs~\cite{kuckreja2024geochat, zhang2024llava, cheng2024spatialrgpt} that target specialized domains and tasks. For instance, recent work has focused on adapting these models for object detection~\cite{zhang2024llava} and depth estimation~\cite{cheng2024spatialrgpt}. While text is the primary modality for LVLMs, it struggles to convey complex or nuanced visual ideas efficiently, more precisely tasks involving fine-grained images. Communicating intricate spatial arrangements or specific object attributes through text alone can be cumbersome and ambiguous. Hand-drawn sketches offer a powerful solution by enabling intuitive visual prompting. Furthermore, sketches transcend linguistic barriers, making them a more accessible and universal communication tool compared to text, which demands descriptive proficiency from the user.

Despite their expressive power, the inherent nature of sketches makes them challenging for machine perception~\cite{mathur2025clipdraw}. They are highly abstract and exhibit significant variability based on artistic style, cultural background, and drawing skill~\cite{alaniz2022abstracting}. This complexity has made sketches a long-standing subject of study in computer vision, driving research in tasks like sketch-based image retrieval (SBIR)~\cite{koley2024handle, koley2024walkalone}, object detection~\cite{chowdhury2023can, tripathi2024query, tripathi2020sketch}, classification~\cite{tiwari2024sketchgpt, bandyopadhyay2024generalised, koley2025sketchfusion}, segmentation~\cite{koley2025sketchyourseg, koley2025freestyle}, and generation ~\cite{liu2025sketchvideo, koley2023picture, banerjee2024svgcraft}. It is this variability and abstraction that current open-source LVLMs struggle to comprehend.

Consider a case where we try to query the existing LVLMs with crude hand-drawn sketches, to give a detailed description of the sketch.  Despite excelling at understanding natural images and structured visual inputs, current models consistently fail to make sense of crude sketches. More importantly, even when they grasp some visual cues, they are unable to leverage this information to perform tasks as shown in Figure ~\ref{fig:motivation}. Such a scenario highlights a critical gap: while sketches reside in the image domain, their abstraction and ambiguity make them fundamentally different.  As LVLMs evolve, understanding sketches alongside text and natural images is increasingly crucial. Yet, sketches remain a major blind spot in open-weight models - primarily due to the lack of a large-scale, diverse, open-source dataset combining all three modalities.

To address the aforementioned gap, we introduce \textbf{\texttt{SketchVCL}}, a comprehensive multi-task dataset of image-sketch-instruction triplets to train LVLMs for four fundamental sketch-based reasoning tasks: \textbf{(a)} object localization, \textbf{(b)} image retrieval, \textbf{(c)} counting, and \textbf{(d)} sketch-aware VQA. We use this dataset to train our model \textbf{\texttt{O3SLM}} - a novel LVLM engineered for sketch-based reasoning. Our experiments show that this model achieves state-of-the-art (SOTA) performance on all the aforementioned tasks against open-weight models. By releasing both the dataset and a high-performing model, we aim to unlock the potential of LVLMs to reason with sketches as fluently as natural images.

Our key contributions are as follows:\\
\noindent 1) We construct \textbf{\texttt{SketchVCL}}, a large-scale, multi-task instruction-tuning dataset composed of ⟨image, sketch, instruction⟩ triplets (refer to Table \ref{tab:dataset_composition}). This is enabled by two key innovations: \textbf{(a)} A novel sketch generation pipeline to generate fine-grained sketches paired with specific object instances within an image. \textbf{(b)} A two-stage training structure. First, a large-scale sketch-alignment stage, developing sketch understanding for an open vocabulary setup. Second, an instruction alignment stage to tune our model to task-specific instructions.\\
\noindent 2) We introduce \textbf{\texttt{O3SLM}}, a unified LVLM designed to handle diverse sketch-based grounding and reasoning tasks within a single framework. \textbf{\texttt{O3SLM}} is fine-tuned on our curated dataset, leveraging the multi-task curriculum to achieve superior sketch-image feature alignment.\\
\noindent 3) We demonstrate through comprehensive evaluation that \textbf{\texttt{O3SLM}} establishes new SOTA performance on sketch-guided object localization, image retrieval, and VQA (counting too) against open-sourced LVLMs. Notably, our single model significantly outperforms general-purpose zero-shot LVLMs, even surpassing closed source models above its weight class like GPT-4o and Gemini 1.5 Pro.

%% file: sec/2_releated_work.tex
\section{2. Related Works}

\paragraph{Sketches as a Visual Modality.} Sketches offer a uniquely abstract and expressive visual modality, widely explored in tasks such as SBIR~\cite{koley2024handle, koley2024walkalone}, object detection~\cite{tripathi2020sketch, tripathi2024query, chowdhury2023can}, image synthesis \cite{koley2024s}, and even video generation \cite{liu2025sketchvideo}. Although rich in expressiveness, sketches present challenges for machine perception due to their high variability, abstraction, and inherent noise. These characteristics often result in limited generalization performance, especially when deployed in open-world settings with novel classes or unseen domains \cite{tripathi2024query}. A major cause of this is the lack of large-scale, open-vocabulary pretraining using sketch modalities—most prior works rely on narrowly scoped datasets and task-specific architectures. Our work addresses this limitation through a unified pretraining and instruction tuning framework that aligns sketches with both natural images and text, enabling scalable sketch understanding across multiple vision-language tasks.

\paragraph{Multimodal Fusion of Sketch and Text.}
Sketch and language provide complementary cues for visual reasoning - while sketches capture spatial and shape-based priors~\cite{koley2024walkalone}, text encodes semantic and contextual information~\cite{chowdhury2023scenetrilogy, koley2025sketchfusion}. Prior works have explored sketch-text fusion for tasks like fine-grained SBIR~\cite{baldrati2023zero, saito2023pic2word} and scene captioning, yet these approaches are often limited to single-task pipelines or rely on handcrafted feature fusion mechanisms. Importantly, existing systems do not operate under open-vocabulary constraints, nor do they support unified pipelines that can generalize across task boundaries. To the best of our knowledge, ours is the first work to demonstrate native support for joint sketch-text queries in tasks such as object detection and counting in LVLMs.

\paragraph{Large Vision-Language Models (LVLMs).} Open-sourced LVLMs such as LLaVA~\cite{liu2023llava}, Qwen-VL2~\cite{wang2024qwen2}, DeepSeek-VL2~\cite{wu2024deepseekvl2}, Pixtral~\cite{agrawal2024pixtral}, and Molmo~\cite{deitke2025molmo} have shown impressive multimodal reasoning capabilities. However, these models fail in the presence of abstract visual inputs such as sketches. MiniGPT-v2~\cite{chen2023minigpt} is one of the few LVLMs spatially trained for grounding tasks; however, it lacks native support for multi-image input, disallowing usage of sketches to query images. On the other hand, closed-source models like GPT-4o~\cite{hurst2024gpt} and Gemini 1.5/2.5~\cite{team2024gemini} show some initial sketch understanding, but their multimodal grounding remains weak. Moreover, inaccessibility and lack of interpretability limit their applicability in open scientific research. Our model, \textbf{\texttt{O3SLM}}, is built to bridge this gap by supporting joint reasoning over sketches, natural images, and text, and demonstrates robust generalization in retrieval, localization, and reasoning tasks, even under zero-shot and OOD settings. Recently,  \cite{lee2024interactive, fu2025iterate} attributed the text-to-image retrieval task to LVLMs, yet to adhere to the sketch understanding capabilities. % However, they plug LVLMs into larger pipelines requiring explicit training. We propose a method which allows SBIR training and inference without changing the 

\paragraph{Sketch generations from Images.} Recent works on sketch generation from images or text explore various approaches. Early GAN-based methods ~\cite{li2019photo} are fast but limited in quality and data diversity. Later methods~\cite{mathur2025clipdraw, vinker2022clipasso, xing2024svgdreamer, vinker2023clipascene} use Bézier curves and VLMs like CLIP~\cite{radford2021learning} for alignment, but produce overly simple sketches with clean backgrounds. Diffusion-based models~\cite{xing2023diffsketcher, arar2025swiftsketch} generate high-quality sketches but are extremely slow. For our large-scale dataset, we adopt the Photo2Sketch~\cite{li2019photo} method, which offers better quality than CLIP-based methods. 
%For LVLM evaluation, we also leverage 4 widely used sketch-image datasets, namely: QuickDraw!, Sketchy, Tu Berlin, and COCO.

\paragraph{Sketch Datasets.} While several popular sketch datasets exist—such as QuickDraw! \cite{jongejan2016quick}, Sketchy \cite{sangkloy2016sketchy}, ShoeV2, QMUL-ChairV2 \cite{yu2016sketch}, TU-Berlin \cite{eitz2012tuberlin}, and SketchyCOCO \cite{gao2020sketchycoco}—none are well-suited for large-scale training of LVLMs. QuickDraw and TU-Berlin consist only of class-wise sketches without paired images. Sketchy, ShoeV2, and QMUL-ChairV2 include image-sketch pairs, but these are limited to simple, single-object scenes designed primarily for the SBIR task. Crucially, none of these datasets include textual descriptions or question-answer pairs, which are essential for training LVLMs. SketchyCOCO provides paired sketches for the COCO dataset, but it is limited to instance-level sketches from only 14 object categories and also lacks associated text. We have summarized the limitations of current sketch datasets in supplementary. Our proposed dataset, \textbf{\texttt{SketchVCL}}, addresses these limitations by incorporating a diverse set of images and sketches from multiple sources, along with rich textual annotations. This makes it a valuable resource for integrating sketches into multimodal LLMs at scale.

%% file: sec/3_dataset.tex
\section{3. \textbf{\texttt{SketchVCL}} Dataset}
\label{sec:dataset}

\begin{figure}[!h]
    \centering
    \includegraphics[width=0.46\textwidth]{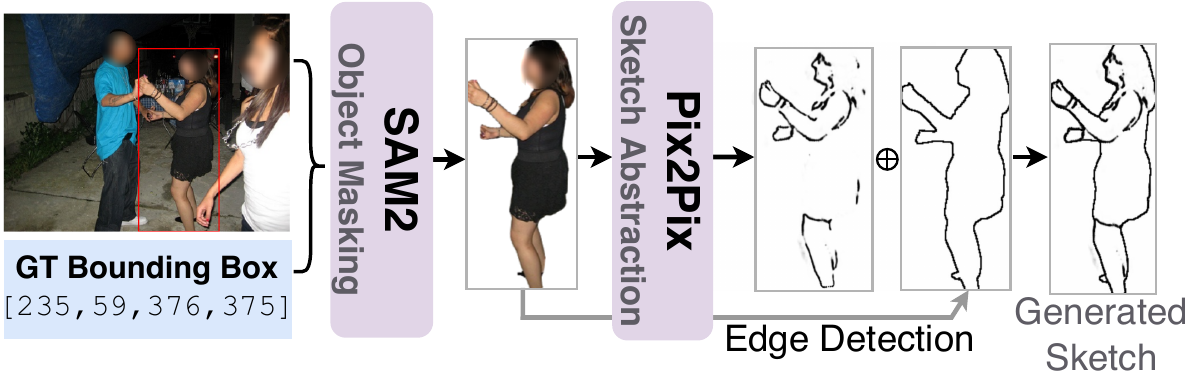}
    \caption{\textbf{Automated Large-Scale Sketch Generation Pipeline.} For each object instance, we use the SAM2-generated segmentation maps to mask the background and pass the foreground through Pix2Pix \cite{li2019photo} for sketch generation. These sketches are enhanced using edge detection using morphological gradients. The final sketch is an aggregation of the edges and the Pix2Pix sketch.}
    \label{fig:curation} 
\end{figure}

Due to the absence of instruction-aligned sketch-image datasets, we propose \textbf{\texttt{SketchVCL}} - a large-scale dataset for sketch-image-text alignment and instruction tuning. \textbf{\texttt{SketchVCL}} comprises 600K instructions for pretraining on OpenImages \cite{kuznetsova2020open} and Object365 \cite{shao2019objects365} datasets, and 50K instructions on COCO \cite{lin2014coco} for finetuning.

OpenImages and Objects365 contain 600 and 365 classes, respectively, which offer broad domain and vocabulary coverage. However, obtaining sketches for such a large class distribution remains a challenge. To address this, we develop a sketch curation pipeline, which generates instance-level sketches for these datasets. This allows us to get paired sketches for large-scale pretraining datasets. Similarly, for finetuning, we generate sketches over COCO to represent the classes absent from Sketchy and QuickDraw!.

\subsection{Sketch Curation Pipeline}

We assume sketches as pixel information abstraction, where the sketches usually consist of contours and edges of the objects. This accumulation of edges will lead to a meaningful yet difficult to comprehend attribute, which is often complementary for explaining objects, as it stores the fine grained details like pose and shape. 

To leverage this, we curate object specific sketch datasets: SketchVCL-OI, SketchVCL-O365, and SketchVCL-C, derived from single object instances within the training subsets of OpenImages, Object365, and COCO, respectively. We have summarized the pipeline in Figure \ref{fig:curation}, through which, we curated 19M and 14M sketches from Object365 and OpenImages, respectively.

% For each instance, we extract object-centric crops using ground-truth bounding boxes, followed by segmenting the object using SAM \cite{kirillov2023segment} to obtain in-the-wild binary masks. These masks isolate the object of interest, which is then passed through the Pix2Pix model \cite{isola2017image} to generate abstracted sketch representations simulating hand-drawn sketches.

% Although due to imperfect abstraction, often due to pixelated, cropped objects and incoherent strokes. To address this, we enhance the edges using the outer contour of the object as shown in the 

\input{tables/dataset}

\subsection{Stage I: Sketch Alignment}

As shown in Table \ref{tab:benchmark_results_count}, current LVLMs struggle to comprehend hand-drawn sketches. To address this issue, we introduce a large-scale pretraining stage. The goal of this stage is to make our model align sketches with the image and text. 

The pretraining phase is designed such to teach the model correspondences across the three modalities: hand-drawn sketches, natural images and text. Specifically, we keep the following goals in mind for the model: i) recognize the sketch, ii) associate the object in the natural image with the sketch, iii) develop fine-grained spatial understanding required for object detection - an ability most LVLMs lack, and iv)  retain its natural language capabilities, such as describing visual content through text. In this way, we ensure 3-way alignment of sketches, images, and text.

To curate our pretraining dataset, we first randomly sample 250K images each from Objects365 and OpenImages. Sketches for these datasets are synthetically generated from our pipeline. For each image, we randomly sample one of the annotated object classes. Further, to ensure adequate representation of all classes, we identify \textit{tail classes} (those with fewer than 5000 instances) and select an additional 50K images, with a balanced sampling strategy over these tail classes, from either datasets.
This results in 600K images, each paired with a target object class. For each image, we generate a descriptive caption of the target objects using DeepSeek-VL2. These captions are further refined and aligned with our task format using LLaMA-3-8B Instruct. Each response begins by identifying the sketch, followed by a detailed description of the object's appearance, its relation to surrounding elements, and concludes with bounding box coordinates around each instance of the target category.

\subsection{Stage II: Instruction Tuning}
While Stage I is responsible for aligning sketches to the images, in Stage II, we align the model to sketch-based tasks and multi-round conversation. Specifically, we train the model across four tasks: Counting, Object Localization, Visual Question Answering (VQA), and Sketch-based Image Retrieval (SBIR). To facilitate instruction tuning, we curate a task-specific dataset, summarized in Table~\ref{tab:dataset_composition}. Following \cite{deitke2025molmo}, we used task specific natural language descriptors as a prefix to the prompt rather than adding tokens to the vocabulary. For all the tasks, the sketches are randomly sampled from SketchMIX for the corresponding class, as described later in this section. Task prompts are also randomly selected to avoid prompt overfitting and improve robustness. Additional implementation details are in the supplementary.

The curated instruction set covers four tasks: \textbf{(1) Counting:} We use the descriptor string \textit{COUNT} as the prefix to the prompt. We use 30K images from the training subset of Pixmo-count dataset, each annotated with ground truth counts. \textbf{(2) Object Detection:} For this task, the descriptor \textit{BBOX} is used as a prefix. We use the training split of COCO to get 110K instructions each with a unique image-class pair. Each instance is an aggregation of all the bounding boxes for a class in an image. \textbf{(3) VQA:} The descriptor string \textit{VQA} is added as a prefix for VQA prompts. Detailed image descriptions are taken from ShareGPT4V \cite{chen2024sharegpt4v}, and Llama-3 \cite{grattafiori2024llama}  generates multi-round, complex reasoning-based questions using the descriptions from ShareGPT4V. To balance sketch-based and general reasoning, we create 25K sketch-based QA pairs and 25K without sketches. \textbf{(4) SBIR:}
We use the descriptor \textit{SBIR} as a prefix to the prompt for this task. Sketchy \cite{sangkloy2016sketchy} images, each containing a single object class are used to curate 12.5K positive and 12.5K negative sketch-image pairs. The answer to this task is strictly \textit{yes} or \textit{no}, which is used to compute accuracy as discussed in Section \ref{sec:results_discussion}5.

\begin{figure}[!t]
    \centering
    \includegraphics[width=\linewidth]{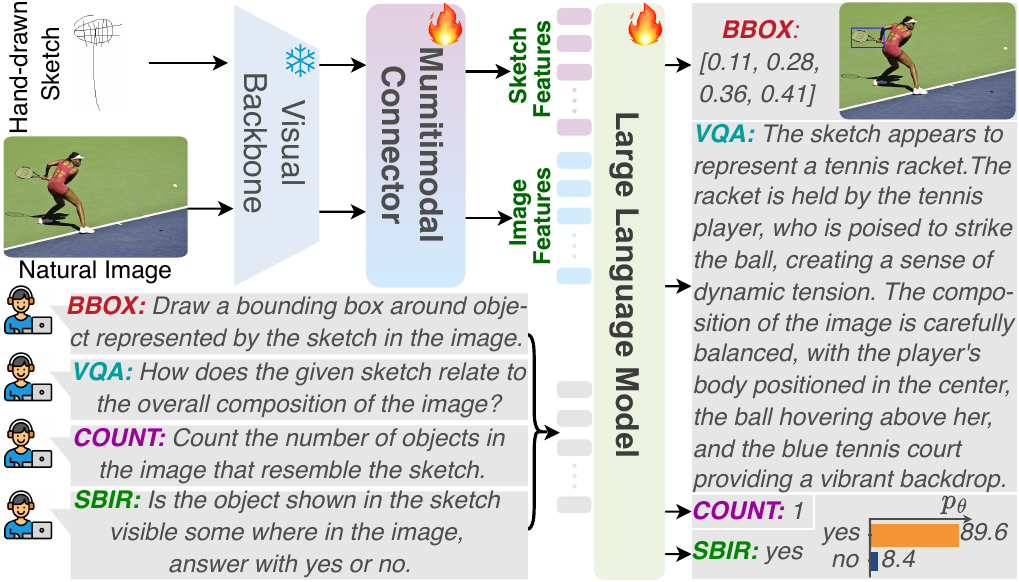}
    
    \caption{\textbf{Summary of \texttt{O3SLM}.} We use CLIP-L-336 as the visual backbone. The hand-drawn sketch and natural image are encoded using this backbone, then the multimodal connector projects the sketch and image features to the input space of the LLM. Finally, the sketch, image, and text tokens are concatenated and passed through the LLM.}
    \label{fig:model_diag}
\end{figure}

\paragraph{SketchMIX.}
For Stage II training, we construct a diverse sketch pool by combining sketches from multiple sources; throughout this paper, we refer to this combination as SketchMIX. To adapt our model to hand-drawn sketches, we include sketches from Sketchy and QuickDraw!, which represent higher quality and abstract sketches, respectively. For more diversity, we also include sketches generated from our pipeline on COCO and Objects365 images. To assess the model’s ability to generalize to an unseen style of sketches, we intentionally exclude the TU-Berlin \cite{eitz2012tuberlin} dataset from SketchMIX. TU-Berlin offers a wide range of abstraction levels in its sketches, which serves well for evaluation. While mixing sketches from various datasets, we need to combine the classes across datasets. Hence, we adopt the Objects365 class taxonomy as a common set and map classes from each of the sketch datasets to this set; we have between 200-350 sketches in every class. We have described the method of curating SketchMIX in detail in the supplementary\footnote{Supplementary material is provided in page 10 along with qualitative results.}.

%% file: tables/dataset.tex
\begin{table}[h!]
\centering
\resizebox{0.48\textwidth}{!}{
\begin{tabular}{clrcr}
% \hline
\textbf{Stage} & \textbf{Task} & \textbf{Image Dataset} & \textbf{Sketch Dataset} & \textbf{\# Size} \\ \shline \multicolumn{4}{c}{\vspace{-10pt}} \\

\multirow{2}{*}{Pretraining} & \multirow{2}{*}{\makecell[l]{Detailed description\\with bounding box}} & Objects365 & SketchVCL-O365 & 300k\\ %\hline
%\cdashlinelr{3-5}
&   & OpenImages & SketchVCL-OI& 300k\\

% \midrule
  \cdashlinelr{1-5}

\multirow{4}{*}{Finetuning} & Object Detection  & COCO & SketchMIX & 110k \\ %\cdashlinelr{2-5}

& VQA & COCO & SketchMIX & 50k \\
% \cdashlinelr{2-5}

& Counting  & PixMo Count & SketchMIX & 30k \\ %\cdashlinelr{2-5}
& SBIR & Sketchy & SketchMIX& 25k \\ %\cdashlinelr{2-4}
\midrule
\end{tabular}}
\caption{\textbf{Training Data Composition.} The distribution of data for each task and corresponding datasets is shown. The total pretraining size is 600k, while the total finetuning size is 215k. Instruction tuning data is curated based on the downstream tasks. More in supplementary.}
\label{tab:dataset_composition}
\end{table}

%% file: sec/4_methods.tex
\input{tables/counting_merge}

\section{4. \textbf{\texttt{O3SLM}}}
\label{sec:method}

Our proposed \textbf{\texttt{O3SLM}} aims to align hand-drawn sketches to the visual information learnt by open LVLMs. We use CLIP ViT-L/336 as the visual backbone to encode both sketches and natural images; the 336px variant allows a higher input resolution for better fine-grained spatial understanding. The extracted embeddings are mapped to the LLM's input space via the multimodal connector (a two-layered MLP). Our base language model is Vicuna v1.5. We initialize the model weights from LLaVA-1.5 to leverage its strong text-image alignment and large-scale training. In summary, our architecture has three components: i) a visual backbone (CLIP-L), ii) a multimodal projector (MLP), and iii) a large language model (LLM). We concatenate the sketch, image, and text tokens before feeding them to the LLM. Since concatenating two sequences and applying self-attention is a more powerful alternative to cross-attention across these sequences, we avoid introducing additional alignment modules. This yields a conceptually simpler yet effective framework, relying on the LLM’s internal self-attention and large capacity to learn alignment implicitly. The architecture is summarized in Figure~\ref{fig:model_diag}.

\paragraph{Tasks.}
We explicitly train \textbf{\texttt{O3SLM}} for four tasks during the fine-tuning stage. For each task, we provide a sketch, an image, and a text prompt. Following Molmo \cite{deitke2025molmo}, we prefix each task with a small string unique to each task. This aligns the model's output to a consistent format, which is helpful during evaluation. We train the model for multi-round visual question answering by prompting the model with an image, a textual question, and a sketch to refer to a specific object. The questions ask the model to describe the object in the image referred to by sketches. Apart from visual attributes like color, appearance, surroundings, etc, we also question the model about things like the purpose of objects, etc. For counting, we ask the model to count how many instances of a sketch exist in the image and train it to return a single integer. For the localization task, we train the model to return bounding boxes in $\{[x_1, y_1, x_2, y_2]\}$ format. We also introduce a simple yet effective approach for image retrieval using LVLMs, which fits into current frameworks for training and inferring LVLMs. We have summarized these tasks in Figure~\ref{fig:main_fig} and discuss them in detail in Section 5. Notably, our model is able to generalize across tasks and is able to utilize fine-grained queries.

%% file: tables/counting_merge.tex
\begin{table*}[!t]
    \centering
    \small
    \setlength\tabcolsep{2.3pt}
    \resizebox{\textwidth}{!}{%
    \begin{tabular}{
        lcccc>{\cellcolor{tablegray}}c|cccc>{\cellcolor{tablegray}}c
    }
    % \specialrule{0pt}{-10pt}{0pt}
    \multicolumn{1}{c}{\textcolor{gray}{}} & \multicolumn{5}{c|}{\textcolor{gray}{\textbf{PixMo-Count}}} & \multicolumn{5}{c}{\textcolor{gray}{\textbf{COCO}}} \\
    % \cmidrule{2-6} \cmidrule{7-11}
    \cmidrule(lr{.75em}){2-6} \cmidrule(lr{.5em}){7-11}

% \multicolumn{1}{c}{\textcolor{gray}{}} & \multicolumn{5}{c}{\textcolor{gray}{\textbf{PixMo-Count}}} & \multicolumn{5}{c}{\textcolor{gray}{\textbf{COCO}}} \\
%     \cmidrule{2-6} \cmidrule{7-11}

    & \textbf{Sketchy} & \textbf{QuickDraw!} & \textbf{Tu Berlin$\textcolor{molmocolor}{^\dagger}$} & \textbf{SketchVCL-C} & \textbf{Avg.}{} & \textbf{Sketchy} & \textbf{QuickDraw!} & \textbf{Tu Berlin\textcolor{molmocolor}{$^\dagger$}} & \textbf{SketchVCL-C} & \textbf{Avg.}{} \\
        
    \textbf{Large Vision-Language Models} & \textcolor{gray}{(153)} & \textcolor{gray}{(194)} & \textcolor{gray}{(453)}  & \textcolor{gray}{(457)}  && \textcolor{gray}{(180)} & \textcolor{gray}{(299)}  & \textcolor{gray}{(430)} & \textcolor{gray}{(472)} & \cellcolor{tablegray}\\ \shline 
    \multicolumn{1}{c}{} & \multicolumn{5}{c|}{\vspace*{-7pt}} & \multicolumn{5}{c}{}\\
    % \\
    % \vline{~----}
    % \hhline{|~|--|}

    % \multicolumn{12}{@{}l}{\vspace{-8pt}}
    % \multicolumn{1}{@{}l}{
    % \makebox[-1pt][l]{\textbf{\textit{API call only}}} \\
    % \vspace*{10pt}
    \hspace*{-4pt}\noindent\textcolor{gray}{\textbf{\textit{API call only}}} &&&&&&&&&& \cellcolor{tablegray} \\

    % \noindent{\textbf{\textit{API call only}}} &&&&&& &&&&& \cellcolor{tablegray} \\

    GPT-4o-mini~\cite{hurst2024gpt} & 48.4 & 40.2 & 20.5 & 17.5 & 31.7 & 14.4 & 19.1 & 11.6 & 14.2 & 14.8\\
    GPT-4o~\cite{hurst2024gpt} & 45.1 & 45.9 & 24.7 & 18.8 & 33.6 & 13.9 & 20.1 & 12.8 & 18.9 & 16.4\\
    Gemini 1.5 Flash~\cite{team2024gemini} & 33.3 & 30.4 & 15.2 & 10.9 & 22.5 & 17.2 & 24.0 & 11.6 & 16.7 & 17.4\\
    Gemini 1.5 Pro~\cite{team2024gemini} & 37.3 & 42.3 & 27.6 & 22.8 & 32.5 & 16.1 & 19.1 & 16.0 & 16.7 & 17.0\\

    \midrule
    % \cdashlinelr{1-12}

    \hspace*{-4pt}{\textcolor{gray}{\textbf{\textit{Open weights ($\approx$ 7B Model Size)}}}} &&&&&&&&&& \cellcolor{tablegray} \\

    LLaVA-1.5-7B~\cite{liu2023llava} & 22.2 & 18.0 & 13.2 & 10.6 & 16.0 & 16.1 & 12.7 & 10.5	& \phantom{2}9.1 & 12.1\\
    
    Qwen2.5-VL-7B~\cite{wang2024qwen2} & 25.5 & 25.3 & 13.3 & \phantom{2}6.8 & 17.7 & \phantom{2}2.2 & \phantom{2}6.0 & \phantom{2}4.2 & \phantom{2}5.9 & \phantom{2}4.6 \\

    DeepSeek-VL2-small~\cite{wu2024deepseekvl2} & 40.5 & 20.6 & 12.4 & \phantom{2}8.1 & 20.4 & \phantom{2}8.9 & \phantom{2}7.4  & \phantom{2}2.8 &	\phantom{2}5.3	& \phantom{2}6.1 \\
    
    Molmo-7B-D~\cite{deitke2025molmo} & 32.7 & 36.6 & 32.7 & 19.3 & 30.3 & 19.4 & 20.4 & \phantom{2}2.3 & \phantom{2}5.9 & 12.0\\

    \textcolor{molmocolor}{O3SLM-7B (Ours)} & {41.8} & {33.0} & {50.6} & {48.6} & \textbf{43.5} & 35.6 & 29.8 & 30.2 & 29.7 & \textbf{31.3}\\
    
    \midrule

    \hspace*{-4pt}{\textcolor{gray}{\textbf{\textit{Open weights ($\approx$ 13B Model Size)}}}} &&&&&&&&&& \cellcolor{tablegray} \\
    % \Xhline{1.2pt}

    LLaVA-1.5-13B~\cite{liu2023llava} & \phantom{2}1.3 & \phantom{2}8.8 & \phantom{2}2.0 & \phantom{2}9.9 & \phantom{2}5.5 & \phantom{2}2.8 & 10.0	& \phantom{2}2.6 & \phantom{2}6.1 & \phantom{2}5.4\\
    
    Pixtral-12B~\cite{agrawal2024pixtral} & 39.2 & 26.8 & 34.4 & 12.7 & 28.3 & 16.7 & 14.0 & 18.4 & \phantom{2}8.1 & 14.3 \\

    DeepSeek-VL2\:~\cite{wu2024deepseekvl2} & 21.6 & \phantom{2}9.8 & \phantom{2}5.5 & \phantom{2}2.6 & \phantom{2}9.9 & \phantom{2}0.6	& \phantom{2}0.7 & \phantom{2}0.0 & \phantom{2}1.3	& \phantom{2}0.6\\
    
    \textcolor{molmocolor}{O3SLM-13B (Ours)} & {45.1} & {37.6} & {47.0} & {46.4} & \textbf{44.0} & 36.7 & 29.8 & 30.5 & 29.9 & \textbf{31.7}\\

    % \bottomrule
    % \hline

    \end{tabular}
      }%
    \caption{\textbf{Evaluation on Sketch-Based Counting.} We evaluate performance on images from COCO and PixMo-Count datasets \cite{deitke2025molmo}. COCO presents a more challenging setting, with a more object categories per image, forcing the model to rely more on the sketches as a query. We sample sketches from four datasets representing various levels of abstraction and difficulty of hand-drawn sketches, for example QuickDraw! has highly abstract and often incomplete sketches. $\textcolor{molmocolor}{\dagger}$ indicates sketch datasets which are unseen by our model during training; they assess our model's ability to generalize to sketch styles.}
    \label{tab:benchmark_results_count}
\end{table*}

%% file: sec/5_experiments.tex
\section{5. Results and Discussion}
\label{sec:results_discussion}

In this section, we present the performance of \texttt{\textbf{O3SLM}} on sketch-based object detection, counting, and image retrieval tasks, followed by ablation studies and analysis.

\subsection{5.1. Performance on Downstream Tasks}
In this section, we evaluate \textbf{\texttt{O3SLM}} across various sketch-based tasks. Further, we demonstrate that \textbf{\texttt{O3SLM}} can handle combined text-sketch queries, showing its fine-grained multimodal understanding.

\paragraph{Counting.}

We introduce the sketch-based object counting task to evaluate a model’s ability to identify and count objects specified solely through a sketch. Unlike full object detection or localization, this task focuses on counting instances, making it a comparatively lighter task. Given the natural image, sketch pair, and ground-truth count of the class: $X_i =(I_i, S_i, c_i)$, the goal is to predict how many instances of the sketched object appear in the image. The natural images $I_i \in D_v$ are sampled from the validation subsets of COCO and Pixmo-count, while $S_i \in D_s$ is sampled from 4 Sketch datasets: QuickDraw, TU Berlin, Sketchy, and SketchVCL-C. The accuracy for the counting task is computed as, $\text{Accuracy} = \frac{1}{N} \sum_{i=1}^{N} \mathbf{1}\left[\, \hat{c}_i = c_i \,\right]$ here, $\hat{c}_i$ represents the prediction for the $i^{th}$ instance. 

As shown in Table \ref{tab:benchmark_results_count}, our model generalizes well to complex counting scenarios, performing strongly on COCO-Count, which averages three object classes per image. Pixmo-Count, with mostly single-class images, is simpler, and the model remains competitive. In a pure zero-shot setting on TU Berlin sketches, it achieves the best performance, demonstrating robust cross-modal transfer.

\paragraph{Object Detection.}
\input{tables/detection}
Object detection with LVLMs is challenging because next-token prediction does not align with spatial localization. Following \cite{kuckreja2024geochat}, we therefore report Accuracy as a softer and more interpretable metric instead of mAP. As shown in Table \ref{tab:benchmark_results_detect}, our model substantially outperforms prior work. The very low accuracy of existing models highlights how introducing sketch queries severely degrades detection performance due to the large domain gap between sketches and natural images.

\paragraph{Sketch-based Image Retrieval.}

\begin{table}[  ]
    \centering
    \begin{tabular}{lccc}
        \textbf{Models} & Acc@1 & Acc@5 & Acc@10\\ \shline \multicolumn{4}{c}{\vspace{-9pt}} \\
        LLaVA-1.5-7B & 11.0 & 14.4 & 13.0 \\
        \cellcolor{tablegray}\textcolor{molmocolor}{O3SLM-7B} & \cellcolor{tablegray}\textbf{65.0} & \cellcolor{tablegray}\textbf{59.2} & \cellcolor{tablegray}\textbf{39.4}\\ \midrule 
        LLaVA-1.5-13B & 10.0 & \phantom{2}9.2 & \phantom{2}8.3\\
        \cellcolor{tablegray}\textcolor{molmocolor}{O3SLM-13B} & \cellcolor{tablegray}\textbf{55.0} & \cellcolor{tablegray}\textbf{46.4} & \cellcolor{tablegray}\textbf{32.9}\\
        \end{tabular}
        \captionof{table}{\textbf{SBIR.} Performance on Sketchy dataset. The substantial gains indicate that although the original LLaVA has very limited sketch understanding, our training data and methodology align sketches and text in \textbf{\texttt{O3SLM}}.}
        \label{tab:sbir_acc}
\end{table}

For a given sketch $S$ and a gallery of images $I = \{I_1, I_2, \cdots I_N\}$, the task of sketch-based image retrieval (SBIR) aims to retrieve the top-k images from the gallery, which align closely with the sketch $S$. Let $T$ represent the text prompt and $X_i = \{I_i, S, T\}$ be an input triplet. We define our training objective as:

\begin{equation}
\begin{split}
\mathop{\rm argmin}\limits_{\theta} \hspace{0.2cm} & -\sum_{i=1}^N \left[ y_i \log (p_\theta(\texttt{<yes>}|X_i)) \right. \\
& \left. + (1-y_i) \log (p_\theta(\texttt{<no>}|X_i)) \right]
\end{split}
\end{equation}

Here, \texttt{<yes>} and \texttt{<no>} are tokens from the LLM's vocabulary. $y_i$ represents the ground truth label; it is 1 for positive classes (i.e., the image $I_i$ matches the sketch $S$) and 0 for negative classes.
This objective is very similar to the binary cross-entropy, and it can directly be used to train LLMs without altering their training framework.

During inference, we simply select the top k images with the highest confidence for the \texttt{<yes>} token. This is achieved by sorting the probabilities in descending order and selecting the top-k entries: $\mathop{\rm argsort}\limits_{i} p_\theta(\texttt{<yes>}|X_i)[-k:]$.

We report SBIR results on the Sketchy dataset in Table~\ref{tab:sbir_acc}. For evaluation, we rank the gallery images based on the probability assigned to the token \texttt{<yes>} that a given sketch corresponds to each of the gallery images. We compute top-$K$ accuracy (Acc@K) by checking how many of the top-$K$ retrieved images belong to the same class as the query sketch.  We compute these metrics across all query sketches. It is important to note that since there are only 5 matching gallery images per query, the maximum achievable Acc@10 is 50. For each sketch, we perform forward passes with all the gallery images, resulting in a total of 10,000 forward passes for the 100 query sketches. Table~\ref{tab:sbir_acc} shows that the baseline LLaVA-1.5 struggles to interpret sketches and associate them with corresponding images. In contrast, \textbf{\texttt{O3SLM}} demonstrates a strong understanding of sketches.

\subsection{5.2. Model Ablations}

\begin{figure}
    \centering
    \includegraphics[width=0.9\linewidth]{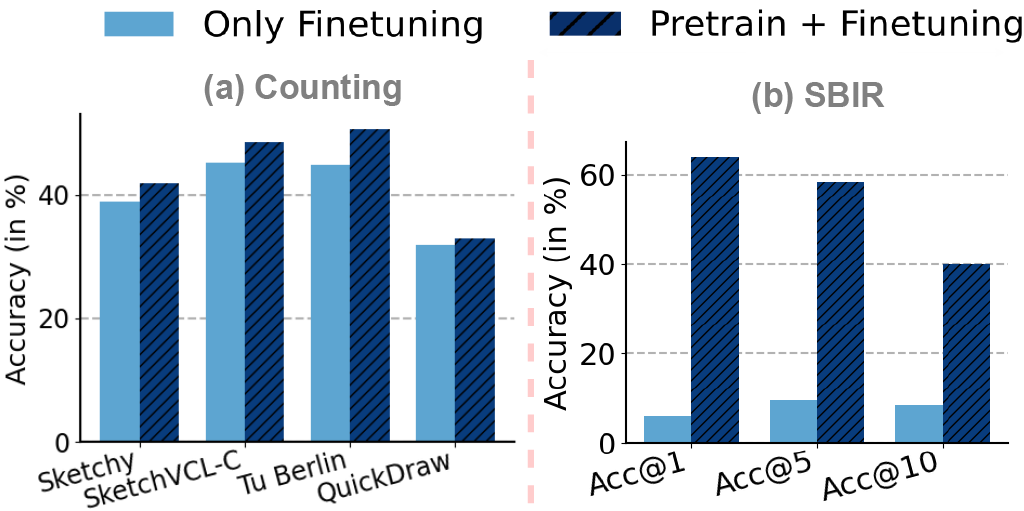}
    \caption{\textbf{Effect of Pretraining.} We assess the impact of our large-scale pretraining stage on two tasks. SBIR tasks significantly benefit from pretraining (Right), whereas the effect on counting is minimal (Left).}
    \label{fig:ptvsft}
\end{figure}

Figure~\ref{fig:ptvsft} discusses the effect of skipping the pretraining stage, being highly sketch dependent SBIR benefits significantly from pretraining. We discuss these results for the detection task in the supplementary.

\begin{figure}
    \centering
    \includegraphics[width=0.85\linewidth]{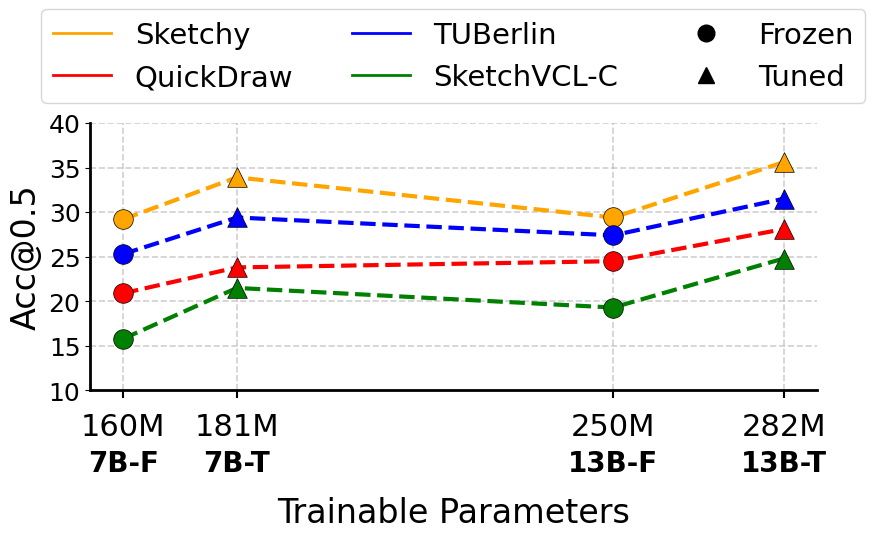}    
    \caption{\textbf{Freezing Multimodal Connector.} We see the trade-off between the performance gain and the extra trainable parameters. We report Acc@0.5 on the detection task.}
    \label{fig:freeze_tune}
\end{figure}

\paragraph{Freezing Multimodal Connector.} As shown by the quantitative metrics in Figure~\ref{fig:freeze_tune}, we get significant gains by tuning the projector along with the LLM. Notably, training only the projector on our 7B model outperforms using a 13B model with a frozen projector. This demonstrates the benefit of aligning sketches and images at the projector level.

\subsection{5.3. Analysis}
\paragraph{Image-Only Performance.}
A natural question to ask is whether \textbf{\texttt{O3SLM}} is still capable of original image only tasks performed by LLaVA. In Table \ref{tab:vis_only}, we see $<5\%$ decrease in popular image-only benchmarks, and we see a large jump in text-only detection compared to LLaVA.
\input{tables/vision_only_apdx}

\paragraph{Emergent Fine-Grained Understanding.}

We qualitatively demonstrate that our model, \textbf{\texttt{O3SLM}}, can perform fine-grained SBIR tasks despite not being explicitly trained to use text descriptions as queries. This capability emerges from extensive multitask pretraining, where the VQA task serves as an auxiliary supervision signal, enhancing the model’s ability to capture fine-grained semantics. By complementing the sketch with a textual description—capturing attributes such as color and texture that are typically difficult to convey through hand-drawn sketches—\textbf{\texttt{O3SLM}} is able to generalize to more nuanced instruction and incorporate textual understanding even in settings it was not explicitly trained for. As shown in Figure~\ref{fig:main_fig}, objects can be queried not only through color or texture but also based on their surroundings or their interaction with other objects. We have included more examples in the supplementary materials.

\paragraph{Visual Question Answering.} We present qualitative analysis on the VQA task in Figure~\ref{fig:main_fig} and the supplementary.

% material. These results demonstrate that \textbf{\texttt{O3SLM}} can effectively incorporate sketches as queries for question answering

%% file: tables/detection.tex
\begin{table*}[!ht]
    \centering
    \small
    
    \setlength\tabcolsep{1.5pt}
    \resizebox{\textwidth}{!}{%
    \begin{tabular}{
        lccccc|ccccc|ccccc|ccccc
    }
    
    \multicolumn{1}{c}{\textbf{}} & \multicolumn{5}{c}{\textbf{Sketchy}} & \multicolumn{5}{c}{\textbf{QuickDraw!}} &
    \multicolumn{5}{c}{\textbf{Tu Berlin$\textcolor{molmocolor}{^\dagger}$}} & \multicolumn{5}{c}{\textbf{SketchVCL-C}} \\

    \textbf{Models} & \textbf{Acc} & \textbf{Acc@0.5} & \textbf{Acc$^S$}  & \textbf{Acc$^M$} & \textbf{Acc$^L$} & \textbf{Acc} & \textbf{Acc@0.5} & \textbf{Acc$^S$}  & \textbf{Acc$^M$} & \textbf{Acc$^L$} & \textbf{Acc} & \textbf{Acc@0.5} & \textbf{Acc$^S$}  & \textbf{Acc$^M$} & \textbf{Acc$^L$} & \textbf{Acc} & \textbf{Acc@0.5} & \textbf{Acc$^S$}  & \textbf{Acc$^M$} & \textbf{Acc$^L$}\\

    \shline \multicolumn{1}{c}{} & \multicolumn{5}{c|}{\vspace*{-8pt}} & \multicolumn{5}{c|}{} & \multicolumn{5}{c|}{} & \multicolumn{5}{c}{} \\

    LLaVA-1.5-7B 
    & \phantom{2}3.5 & \phantom{2}9.1 & 0.0 & \phantom{2}0.0 & \phantom{2}3.5 
    & \phantom{2}2.7 & \phantom{2}6.9 & 0.0 & \phantom{2}0.2 & \phantom{2}2.8
    & \phantom{2}3.5 & \phantom{2}9.7 & 0.0 & \phantom{2}0.6 & \phantom{2}3.5
    & \phantom{2}2.9 & \phantom{2}7.4 & 0.0	& \phantom{2}1.1 & \phantom{2}2.9

    \\
    OneVision
    & \phantom{2}2.8 & \phantom{2}9.2 & 0.0 & \phantom{2}2.2 & \phantom{2}2.8
    & \phantom{2}3.8 & \phantom{2}9.5 & 0.1 & \phantom{2}0.9 & \phantom{2}3.9
    & \phantom{2}2.7 & \phantom{2}9.4 & 0.0 & \phantom{2}1.8 & \phantom{2}2.8
    & \phantom{2}3.6 & \phantom{2}8.9 & 0.0 & \phantom{2}1.1 & \phantom{2}3.7
    \\
    
    DeepSeek-VL2-small
    & \phantom{2}3.3 & \phantom{2}9.7 & 0.0 & \phantom{2}0.5 & \phantom{2}3.3
    & \phantom{2}3.8 &           10.3 & 0.0 & \phantom{2}1.1 & \phantom{2}3.9
    & \phantom{2}3.8 &           11.4 & 0.0 & \phantom{2}2.2 & \phantom{2}3.8
    & \phantom{2}4.5 &           11.3 & 0.0 & \phantom{2}0.6 & \phantom{2}4.7
    \\

    Molmo-7B-D 
    & \phantom{2}1.8 & \phantom{2}5.3 & 0.0 & \phantom{2}0.7 & \phantom{2}1.8
    & \phantom{2}2.9 & \phantom{2}7.9 & 0.0	& \phantom{2}2.1 & \phantom{2}2.9
    & \phantom{2}2.1 & \phantom{2}7.5 & 0.0	& \phantom{2}2.4 & \phantom{2}2.0
    & \phantom{2}2.1 & \phantom{2}5.3 & 0.0	& \phantom{2}2.0 & \phantom{2}2.0
    \\

    \cellcolor{tablegray}\textcolor{molmocolor}{O3SLM-7B (Ours)} 
    & \cellcolor{tablegray}19.4 & \cellcolor{tablegray}33.9 & \cellcolor{tablegray}2.1 & \cellcolor{tablegray}13.3 & \cellcolor{tablegray}33.2
    & \cellcolor{tablegray}13.6 & \cellcolor{tablegray}23.8 & \cellcolor{tablegray}1.3 & \cellcolor{tablegray}\phantom{2}9.3 & \cellcolor{tablegray}26.1
    & \cellcolor{tablegray}16.8 & \cellcolor{tablegray}29.4 & \cellcolor{tablegray}1.7 & \cellcolor{tablegray}11.0 & \cellcolor{tablegray}29.9
    & \cellcolor{tablegray}11.8 & \cellcolor{tablegray}21.5 & \cellcolor{tablegray}1.3 & \cellcolor{tablegray}\phantom{2}8.5 & \cellcolor{tablegray}23.9
    \\
    \midrule
    
    LLaVA-1.5-13B
    & \phantom{2}4.2 & 11.4 & 0.0 & \phantom{2}0.3 & \phantom{2}4.2 
    & \phantom{2}2.9 & \phantom{2}7.1 & 0.0 & \phantom{2}0.6 & \phantom{2}2.9
    & \phantom{2}3.6 & 10.1 & 0.5 & \phantom{2}0.2 & \phantom{2}3.6
    & \phantom{2}2.8 & \phantom{2}7.4 & 0.3 & \phantom{2}0.9 & \phantom{2}2.9
    
    \\

    DeepSeek-VL2 
    & \phantom{2}2.3 & \phantom{2}6.0 & 0.0 & \phantom{2}0.0 & \phantom{2}2.3
    & \phantom{2}1.7 & \phantom{2}5.6 & 0.0 & \phantom{2}1.2 & \phantom{2}1.8
    & \phantom{2}2.1 & \phantom{2}6.0 & 0.0 & \phantom{2}2.3 & \phantom{2}2.1
    & \phantom{2}1.5 & \phantom{2}4.0 & 0.0 & \phantom{2}0.9	& \phantom{2}1.5
    
    \\

    \cellcolor{tablegray}\textcolor{molmocolor}{O3SLM-13B (Ours)}
    & \cellcolor{tablegray}21.3 & \cellcolor{tablegray}35.6 & \cellcolor{tablegray}2.8 & \cellcolor{tablegray}15.5 & \cellcolor{tablegray}35.2
    & \cellcolor{tablegray}16.7 & \cellcolor{tablegray}28.1 & \cellcolor{tablegray}2.0 & \cellcolor{tablegray}11.8 & \cellcolor{tablegray}29.1
    & \cellcolor{tablegray}18.7 & \cellcolor{tablegray}31.5 & \cellcolor{tablegray}2.1 & \cellcolor{tablegray}13.2 & \cellcolor{tablegray}32.1
    & \cellcolor{tablegray}14.6 & \cellcolor{tablegray}24.8 & \cellcolor{tablegray}1.3 & \cellcolor{tablegray}11.0 & \cellcolor{tablegray}26.0
    \\
    
    % \bottomrule

    \end{tabular}
      }%
    \caption{\textbf{Sketch-Based Object Detection.} To evaluate the sketch-based object detection on images COCO val2017, and sketches from four different datasets, specifically: Sketchy, QuickDraw!, TU-Berlin, and SketchVCL-C. Following \cite{kuckreja2024geochat}, we report the Acc metric, we include mAP scores of our model in our supplementary. $\textcolor{molmocolor}{\dagger}$ indicates sketch datasets that are unseen by our model during training; they assess our model's ability to generalize to sketch styles.}
    \label{tab:benchmark_results_detect}
\end{table*}

%% file: tables/vision_only_apdx.tex
\begin{table}[!ht]
    \centering
    \small
    \setlength\tabcolsep{3pt} % tighter columns
    \begin{tabular}{l|ccccc}
        % \hline
        \textbf{Models} & \textbf{VQA} & \textbf{MME} & \textbf{LLaVA in} & \textbf{SeedBench} & \textbf{Text-only} \\
         \multicolumn{1}{c|}{(13B)} & v2  &  & \textbf{the Wild} & (Image)& \textbf{Obj Det} \\
        \hline
        LLaVA & 80.0 & 1531 & 52.0 & 68.2 & 13.4 \\
        \cellcolor{tablegray}\textcolor{molmocolor}{O3SLM} & \cellcolor{tablegray}76.6 & \cellcolor{tablegray}1414 & \cellcolor{tablegray}48.5 & \cellcolor{tablegray}65.5 & \cellcolor{tablegray}21.0 \\
        % \hline
    \end{tabular}
    \caption{\textbf{Image-only tasks.} Comparison with our baseline LLaVA-v1.5 on the following image only bechmarks: VQAv2~\cite{goyal2017making}, MME~\cite{fu2025mme}, LLaVA-in-the-Wild \cite{liu2023llava}, SeedBench~\cite{li2023seed}. We report Acc for text-only object detection. Higher is better.} 
    \label{tab:vis_only}
\end{table}

%% file: sec/6_conclusion.tex
\section{6. Conclusion}

In this work, we introduced the novel task of sketch understanding in Large Vision-Language models. To this end, we proposed a novel sketch-based pretraining and visual instruction tuning dataset to incorporate sketch-to-image alignment. This was created using the automated sketch generation pipeline, enabling scalable and diverse training samples. Furthermore, we proposed our model trained on generated data, and through extensive evaluations on tasks like count, SBIR, and detection, we showed the effectiveness of our approach. Our model consistently outperforms existing LVLMs, underscoring the value of sketch-aware training.

%% file: sec/acknowledgements.tex
\section{Acknowledgments}

We gratefully acknowledge Kotak-IISc AI/ML Centre (KIAC) for the generous conference travel grant and the GPU resources that enabled this research.

%% file: sec/append.tex
This supplementary is organized as follows.
Section A provides implementation details and the hyperparameters used to train \textbf{\texttt{O3SLM}}.
Section B describes the \textbf{\texttt{SketchVCL}} curation procedure and presents sample data.
Section C reports additional qualitative results and other analyses and results not included in the main paper.
Finally, Section D discusses the limitations of \textbf{\texttt{O3SLM}}.

\section{A. Implementation Details}

\subsection{A.1. Training Settings}
We train \textbf{\texttt{O3SLM}} on two NVIDIA H100 GPUs (95 GB). Both stages use Low rank approximation for memory efficiency. All other optimization settings follow the official LLaVA configuration. We have summarised it below in Table \ref{tab:hparmas}

\begin{table}[h]
    \centering
    \begin{tabular}{c|c}
        \textbf{Hyperparameter} &  \textbf{Value}\\ \hline
        Learning Rate & $2 \times 10^{-5}$ \\ 
        LR Schedule & Cosine Decay \\
        LR Warmup & 3 \% \\
        Weight Decay & 0 \\
        Batch Size & 24 \\
        LoRA rank & 64 \\
        Train Epochs & 1 \\
    \end{tabular}
    \caption{Hyperparamters for Training \textbf{\texttt{O3SLM}}}
    \label{tab:hparmas}
\end{table}

\input{tables/dataset_details_appendix}

\subsection{A.2. Evaluation on SBIR Task}

To ensure fair evaluation, we first sort all classes in the Sketchy dataset based on their object detection mAP (as predicted by our model), and then uniformly sample 20 classes from this sorted list. For each of these 20 classes, we randomly select 5 images and 5 sketches, resulting in a gallery of 100 images and a query set of 100 sketches. Further, we compute the  probability that a given sketch corresponds to each of the 100 gallery images, using the probability assigned to the token \texttt{<yes>}.  The gallery images are then ranked based on this probability. We compute top-$K$ accuracy (Acc@K) by checking how many of the top-$K$ retrieved images belong to the same class as the query sketch.  We compute these metrics across all 100 query sketches. Specifically, Acc@1 is 1 if the top-ranked image matches the sketch class, and 0 otherwise. Acc@5 is computed as the fraction of top-5 images that belong to the same class. For example, if 3 out of the top 5 images are correct, Acc@5 is 0.6 for that sketch.

\section{B. Dataset Details}

\begin{figure*}[!h]
    \centering
    \includegraphics[width=\linewidth]{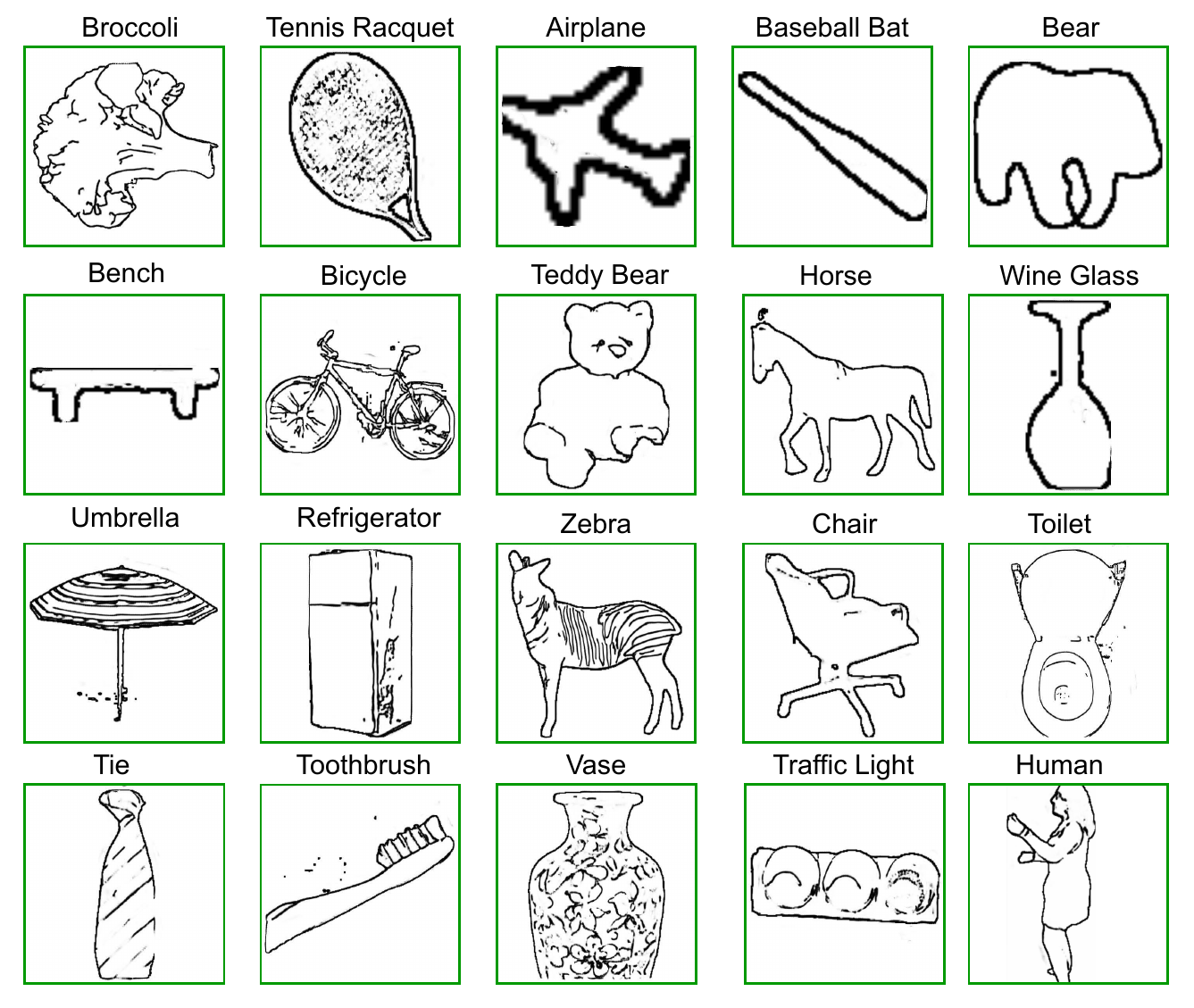}
    \caption{\textbf{Samples from SketchVCL.} We show some qualitative results from our sketch-generation pipeline. We use sketches solely from our pipeline during the pretraining stage. }
    \label{fig:sketchvcl}
\end{figure*}

Table \ref{tab:sketch_dataset_comparison} shows the differences between existing object-centric sketch datasets and SketchVCL. Figure \ref{fig:sketchvcl} shows some class-wise sketches from the SketchVCL dataset.

\subsection{B.1. SketchMIX}

The SketchMIX dataset is a mixture of sketches across  SketchVCL-O365, SketchVCL-C, QuickDraw!~\cite{jongejan2016quick}, and Sketchy~\cite{sangkloy2016sketchy}. 
To create a unified set of object categories across these datasets, we first mapped the classes from Sketchy, QuickDraw!, and SketchVCL-C to the class taxonomy from Objects365~\cite{shao2019objects365}. This mapping was done using CLIP embeddings of class names to identify semantically similar categories across datasets, resulting in a set of parent classes shared between them.

To enhance the diversity of sketches for each object class, we adopt a balanced sampling strategy that favors underrepresented datasets. For each class, we ensure a minimum of 200 sketches. The following rules summarize our curation strategy:

\begin{itemize}
    \item If a class is exclusive to SketchVCL-O365 (i.e., no matching class exists in the other datasets), we sample all 200 sketches from SketchVCL-O365.
    \item  If the class exists in one or more additional datasets, we sample 50 sketches from SketchVCL-O365 and 150 sketches distributed among the other datasets containing that class.
\end{itemize}

This strategy ensures broader visual diversity and reduces dataset bias while maintaining class consistency across SketchMIX.

\begin{figure}[!h]
    \centering
    \begin{minipage}[t]{0.48\linewidth}
        \centering
        \includegraphics[width=\linewidth]{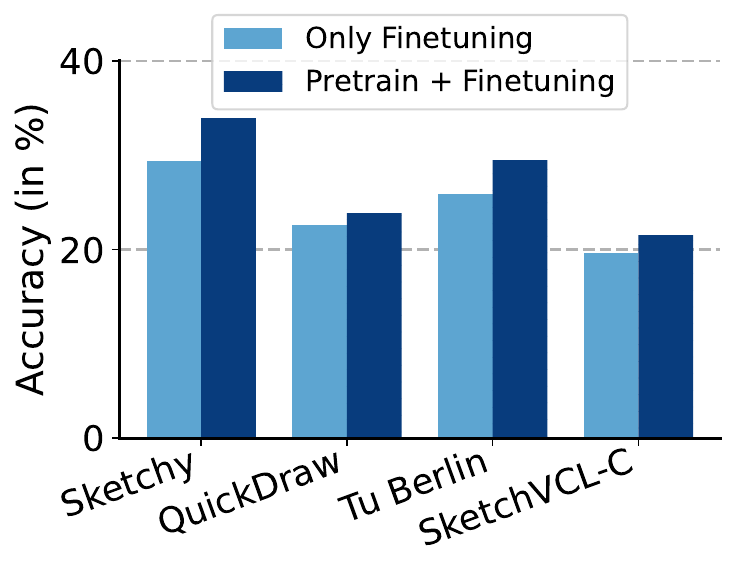}
    \end{minipage}\hfill
    \begin{minipage}[t]{0.48\linewidth}
        \centering
        \includegraphics[width=\linewidth]{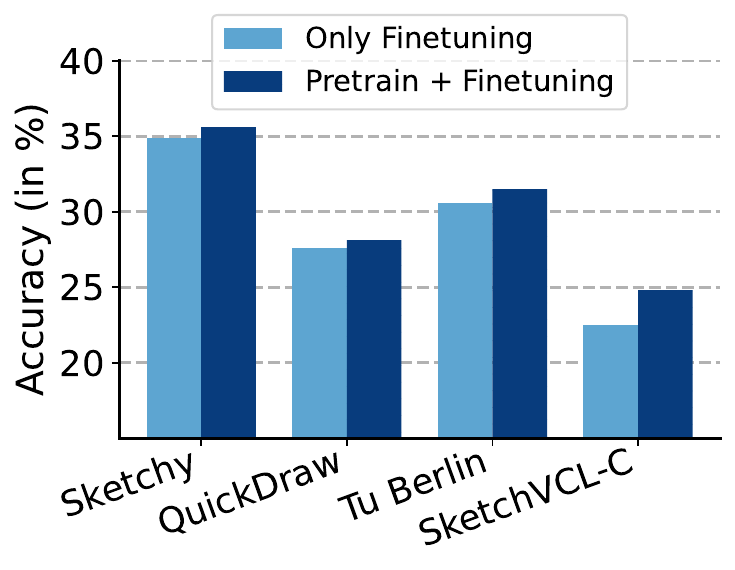}
    \end{minipage}
    \caption{\textbf{Effect of Pretraining.} We compare model performance with and without our large-scale pretraining stage on the Sketch-based Localization task. We report Acc@50 on \textbf{\texttt{O3SLM}-7B} (Left) and \textbf{\texttt{O3SLM-13B}} (Right).}
    \label{fig:ft-pt}
\end{figure}

\subsection{B.2. Format for Pretraining and Instruction Tuning Data generation}

For generating instruction-tuning data, responses for different tasks are generated using DeepSeekVL2 \cite{wu2024deepseekvl2}, by annotating the input image with bounding boxes and asking the model to describe the objects marked with them. Further, the original bounding boxes are modified to lie in [0,1], as in \cite{kuckreja2024geochat}, and either added to the generated text (detection task) or the bounding box count is added (counting task). For pretraining data, we use the same format as VQA by describing the entire scene with respect to the given sketch, and further we append the bounding boxes of the specific object as shown in the Figure \ref{fig:instruction}. Note that all the instances in the data are generated using a single sketch class.

\input{tables/detection_appendix}

\begin{figure}
    \centering
    \includegraphics[width=\linewidth]{figs/supp/data1_down.pdf}
    \caption{\textbf{Examples from the pretraining and  instruction-following data}}
    \label{fig:instruction}
\end{figure}

\begin{figure}[!h]
    \centering
    \includegraphics[width=0.7\linewidth]{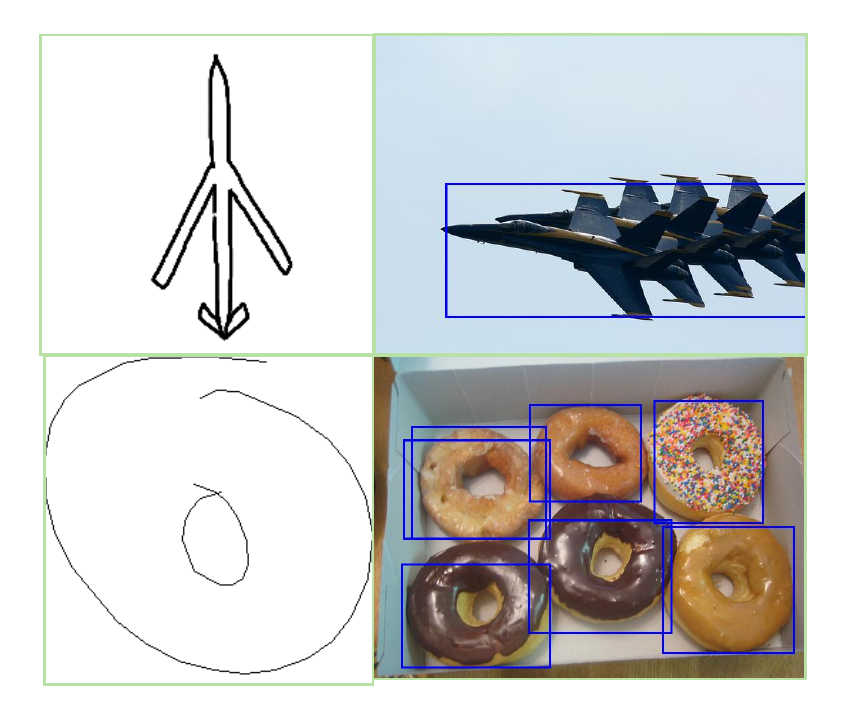}
    \caption{\textbf{Limitations of \textbf{\texttt{O3SLM}}.}}
    \label{fig:limitations}
\end{figure}

\section{C. Additional Results}

\subsection{C.1. Effect of Pretraining}

Figure \ref{fig:ft-pt} shows the need for large-scale pretraining to learn better image and sketch alignment. Pretraining provides a consistent performance boost, especially when the target sketch dataset shares characteristics with the pretraining domain. This supports the importance of initializing models with relevant knowledge before fine-tuning, especially for tasks with sparse or abstract visual input like sketches.

\subsection{C.2. Results on Sketch-based Object Detection}
In Table \ref{tab:detect_map}, we show mAP scores for our model and some other LVLMs. We include these in addition to the Acc scores in the main paper for completeness.

\subsection{C.3. Qualitative Results}

In this section we show some of the qualitative results of our 13B-Tuned model over a) Object Detection b) VQA c) SBIR, with some fine-grained examples.

\begin{figure*}[!h]
    \centering
    \includegraphics[width=0.8\linewidth]{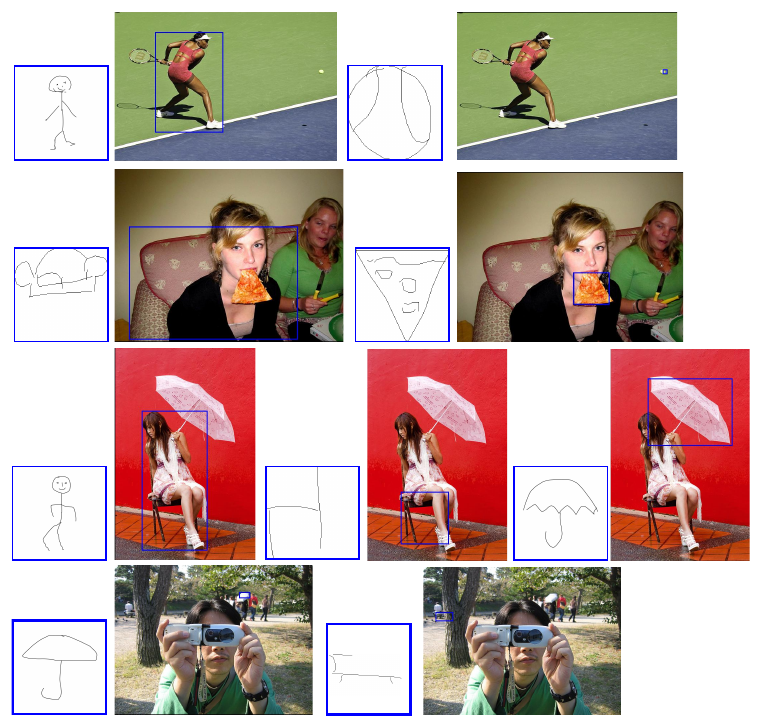}
    \caption{\textbf{Qualitative Results for Detection.} }
    \label{fig:detection_1}
\end{figure*}

\begin{figure*}[!h]
    \centering
    \includegraphics[width=0.8\linewidth]{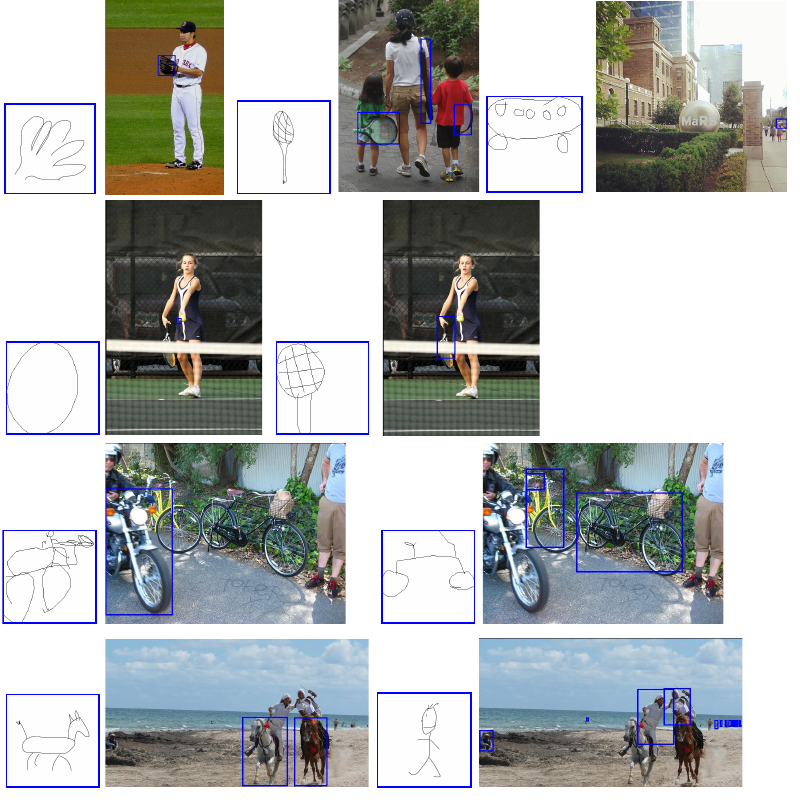}
    \caption{\textbf{Qualitative Results for Detection.} }
    \label{fig:detection_2}
\end{figure*}

\begin{figure*}[!h]
    \centering
    \includegraphics[width=0.8\linewidth]{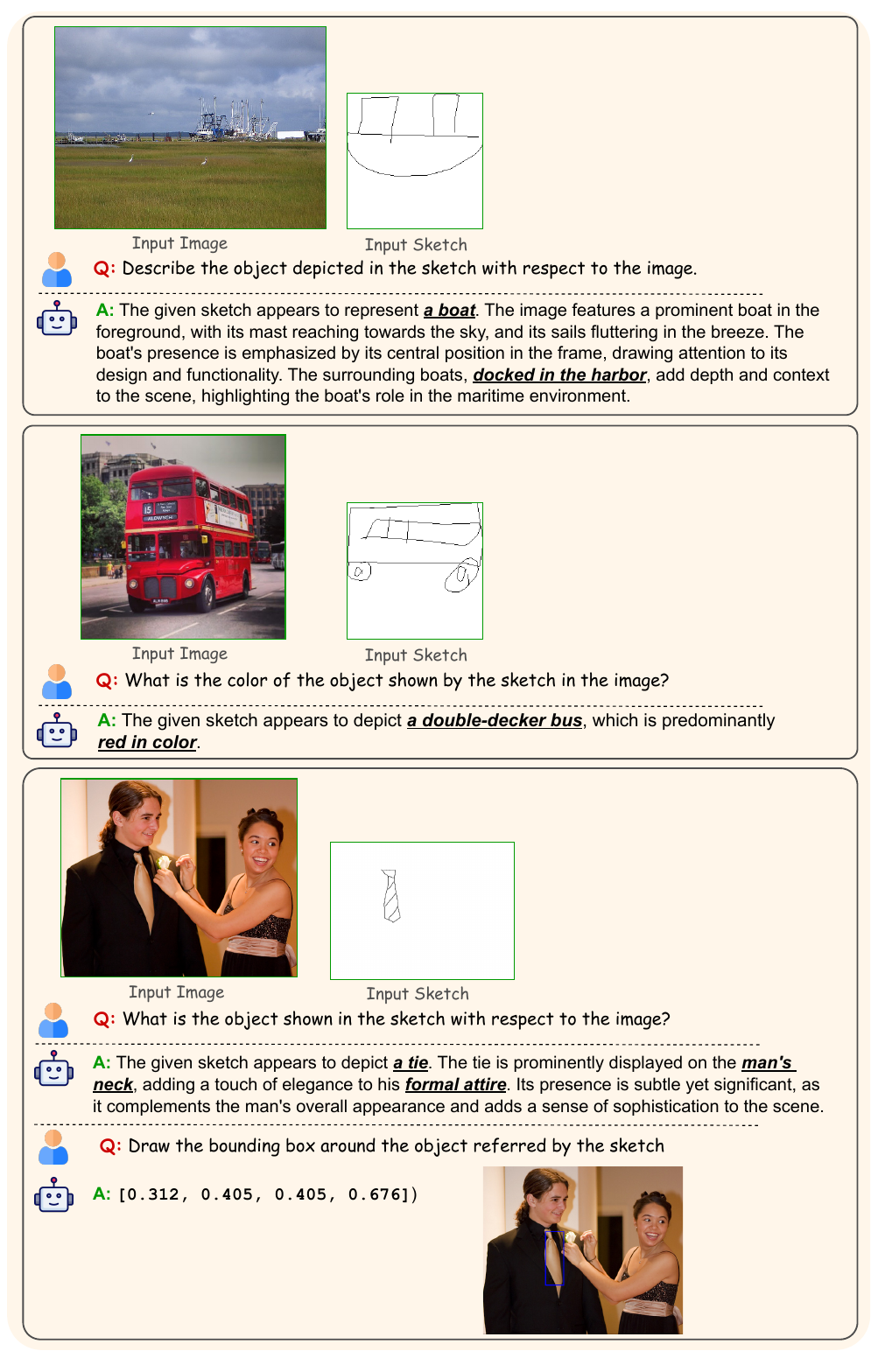}
    \caption{\textbf{Qualitative Results for VQA.} }
    \label{fig:vqa_1}
\end{figure*}

\begin{figure*}[!h]
    \centering
    \includegraphics[width=0.8\linewidth]{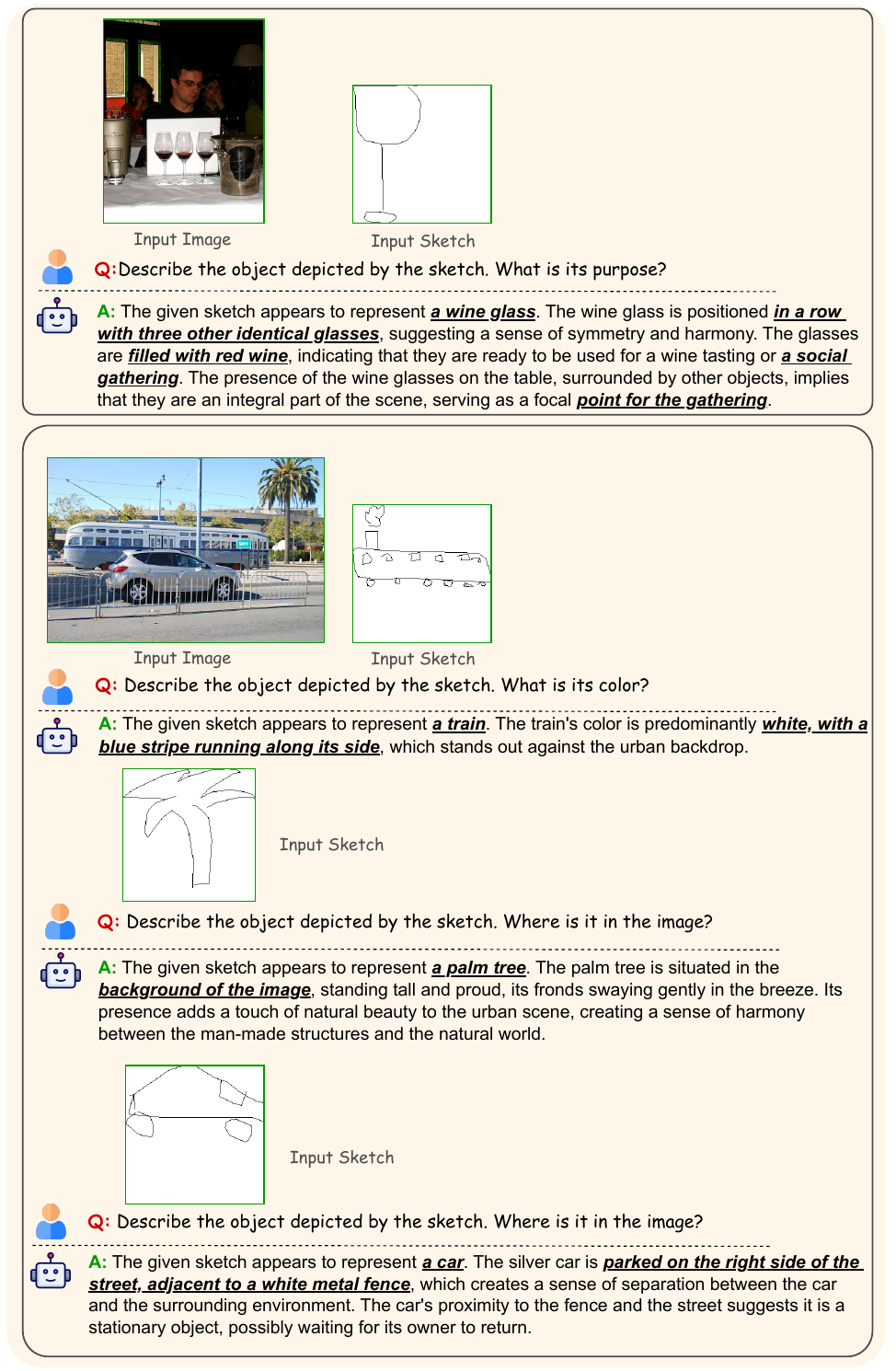}
    \caption{\textbf{Qualitative Results for VQA.} }
    \label{fig:vqa_2}
\end{figure*}

\begin{figure*}[!h]
    \centering
    \includegraphics[width=0.8\linewidth]{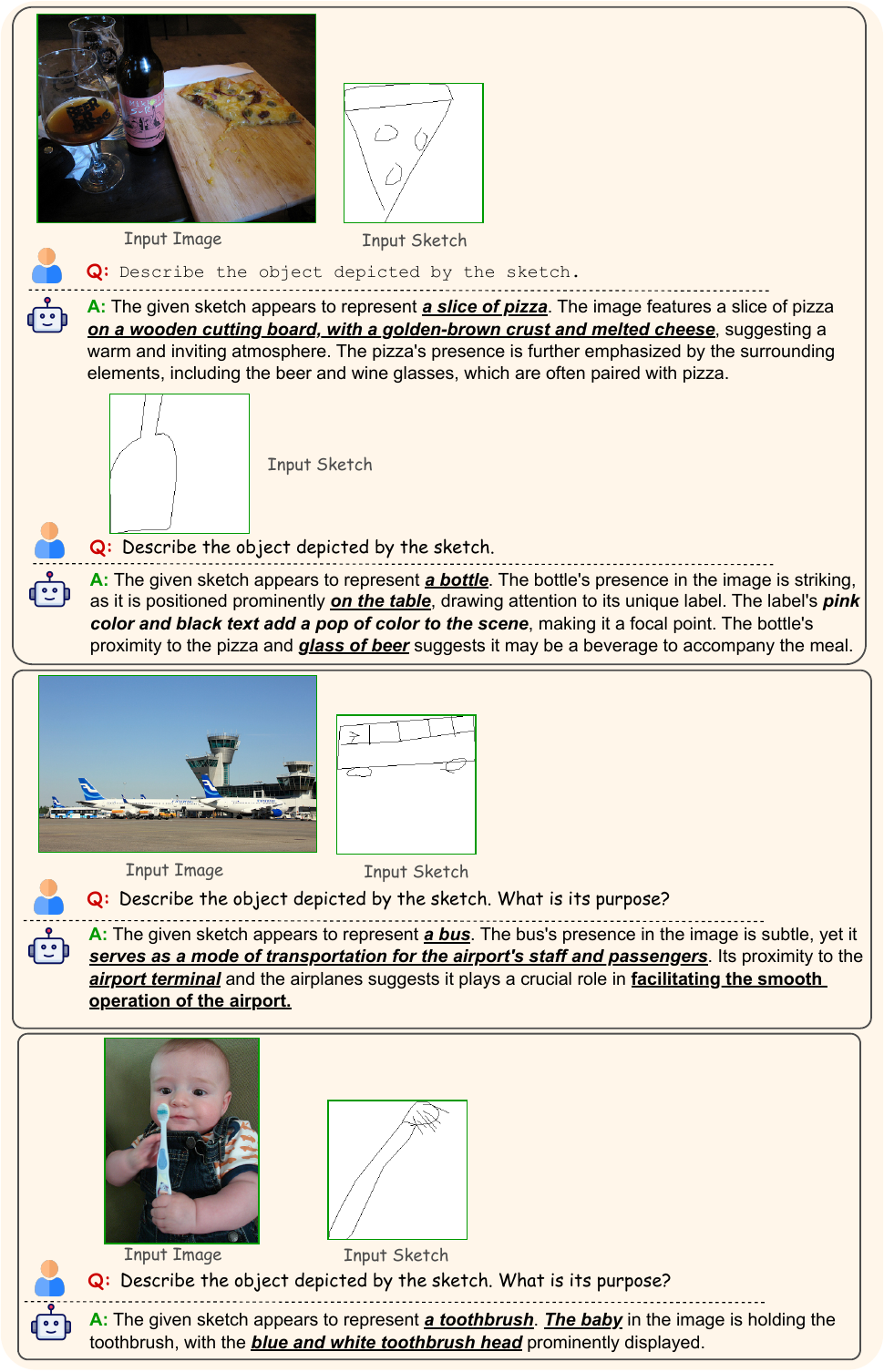}
    \caption{\textbf{Qualitative Results for VQA.} }
    \label{fig:vqa_3}
\end{figure*}

\begin{figure*}[!h]
    \centering
    \includegraphics[width=\linewidth]{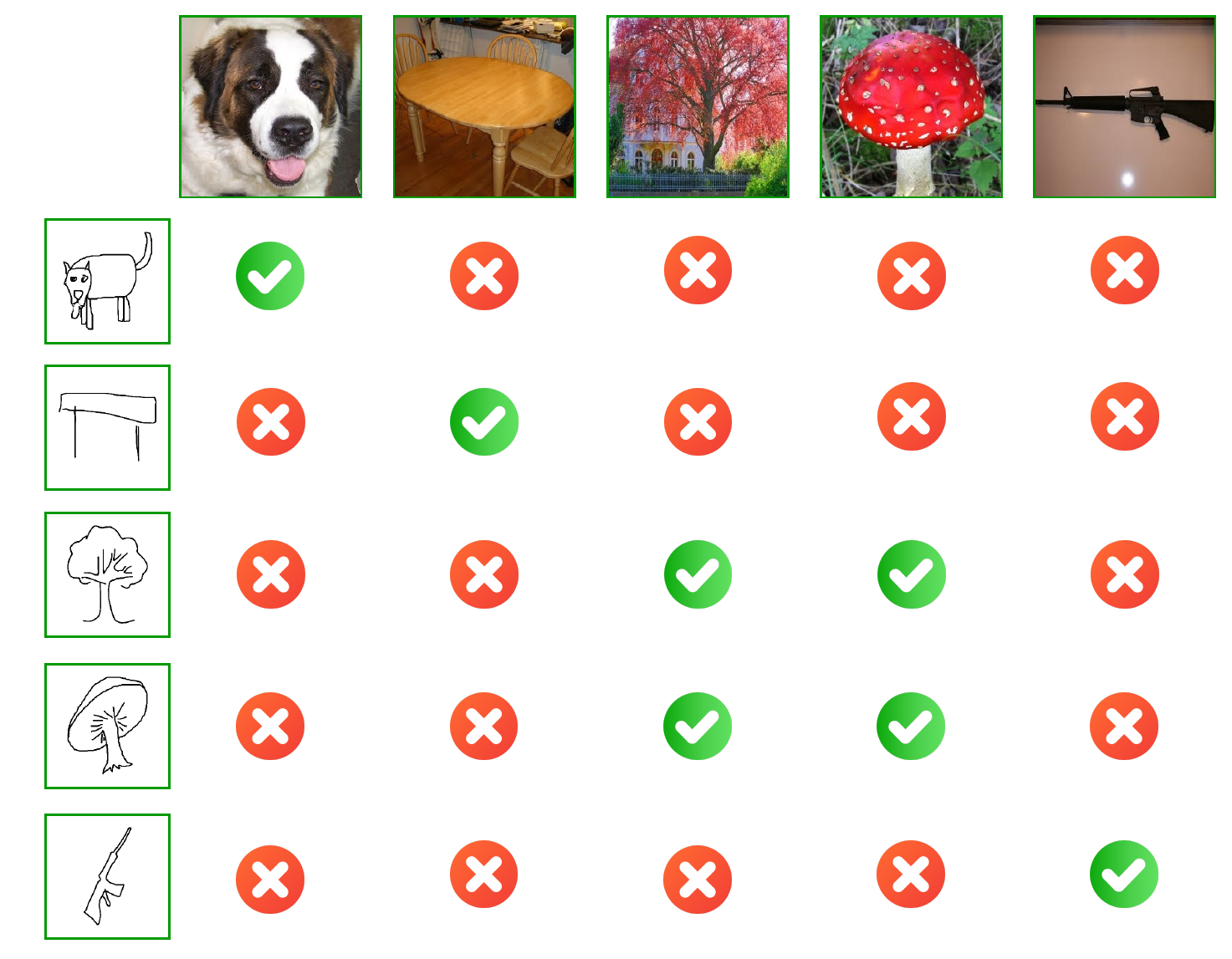}
    \caption{\textbf{Qualitative Results for SBIR.} We show qualitative results for the Sketch-based Image Retrieval Task for \textbf{\texttt{O3SLM}}. We see that \textbf{\texttt{O3SLM}} confuses structurally similar sketches even if the classes are semantically different.}
    \label{fig:sbir_general}
\end{figure*}

\begin{figure*}[h!]
\centering
\begin{subfigure}[t]{\textwidth}
\centering
\includegraphics[width=\textwidth]{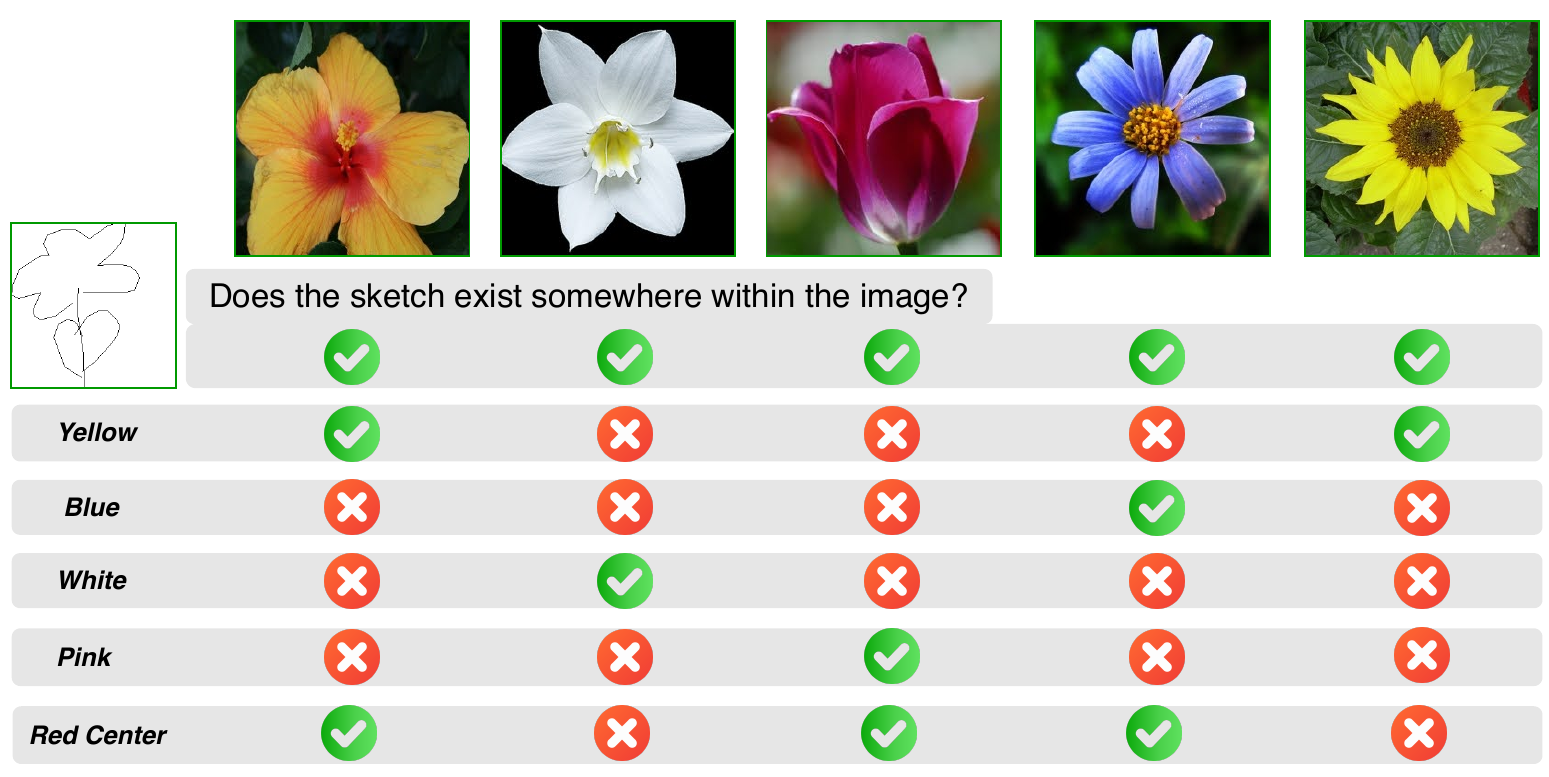}
\label{fig:subim1}
\end{subfigure}
\par\bigskip % Adds a vertical space between the subfigures
\begin{subfigure}[t]{\textwidth}
\centering
\includegraphics[width=\textwidth]{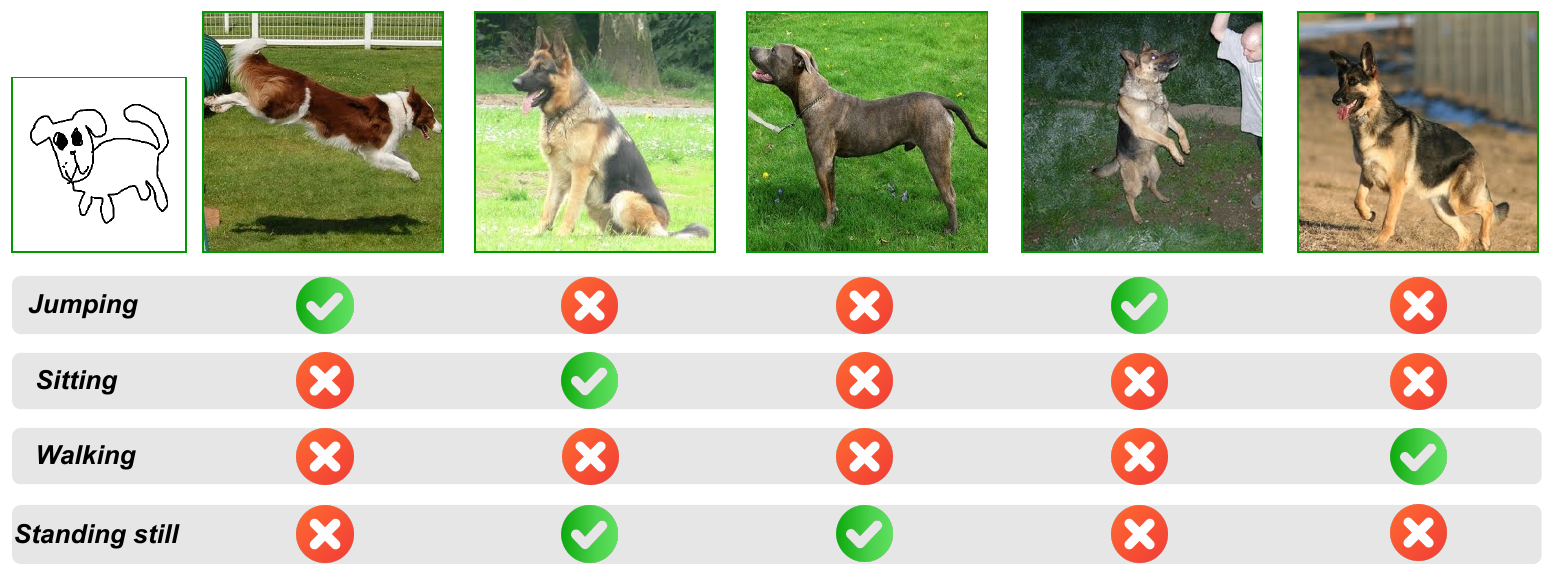}
\label{fig:subim2}
\end{subfigure}
\caption{\textbf{Qualitative Results for FG-SBIR.} Here we show that \textbf{\texttt{O3SLM}} can leverage details from the text query in addition to the hand-drawn sketch for Fine-Grained Sketch-Based Image Retrieval (FG-SBIR). In the figure above, we show examples of \textbf{\texttt{O3SLM}} understanding actions and colors through text prompts for finer control over the image retrieval process.}
\label{fig:example}
\end{figure*}

\section{D. Limitations}
Figure \ref{fig:limitations} shows some of the limitations of model. It struggles in presence of many objects, especially if they have high overlap. In such cases it sometimes, generates extra bounding boxes. Traditional detection algorithms use non-maximal suppression to deal with this issue.

%% file: tables/dataset_details_appendix.tex
\begin{table*}[!t]
\centering
\small
\resizebox{\textwidth}{!}{%
\begin{tabular}{lccccc}
% \hline
\textbf{Sketch Datasets} & \textbf{Automated Pipeline} & \textbf{Level of Abstraction} & \textbf{\# Sketches(instance-level)} & \textbf{\# Images} & \textbf{\# Instructions} \\
% \hline
\shline \multicolumn{6}{c}{\vspace{-8.5pt}}\\

ShoeV2 \cite{yu2016sketch}& \xmark  & Low     & 6.7K & 2K & \xmark  \\
QMUL ChairV2 \cite{yu2016sketch} & \xmark  & Low & 400 & 1.8K & \xmark  \\
TU-Berlin \cite{eitz2012tuberlin} & \xmark  & Low & 20K & \xmark  & \xmark  \\
Sketchy \cite{sangkloy2016sketchy}& \xmark  & Low-Mid & 75K & 12.5K & \xmark  \\
QuickDraw \cite{jongejan2016quick}& \xmark  & Mid-High & 50M & \xmark  & \xmark  \\
SketchyCOCO \cite{gao2020sketchycoco} & \xmark  & Mid & 20K  & 20K & \xmark  \\
ImageNet Sketch \cite{wang2019learning} & \xmark  & Very Low & 50K & \xmark  & \xmark  \\
\hline

\cellcolor{tablegray}\textbf{SketchVCL} & \cellcolor{tablegray}\cmark & \cellcolor{tablegray}\textbf{High} & \cellcolor{tablegray}32M & \cellcolor{tablegray}\textbf{3.7M} & \cellcolor{tablegray}\textbf{650K} \\
\hline
\end{tabular}%
}
\caption{\textbf{Comparison of open-source sketch datasets.} This table shows the comparison between existing sketch-based datasets, which are curated manually (hand-drawn) and our dataset. SketchVCL is the only dataset that has natural language instructions paired with sketches and images. The number of sketches can be further expandable using our novel automated pipeline, which provides sketches with a high level of abstraction.  }
\label{tab:sketch_dataset_comparison}
\end{table*}

%% file: tables/detection_appendix.tex
\begin{table*}[!ht]
    \centering
    \small
    
    \setlength\tabcolsep{1.5pt}
    \resizebox{\textwidth}{!}{%
    \begin{tabular}{
        lccccc|ccccc|ccccc|ccccc
    }
    
    \multicolumn{1}{c}{\textbf{}} & \multicolumn{5}{c}{\textbf{Sketchy}} & \multicolumn{5}{c}{\textbf{QuickDraw!}} &
    \multicolumn{5}{c}{\textbf{Tu Berlin$\textcolor{molmocolor}{^\dagger}$}} & \multicolumn{5}{c}{\textbf{SketchVCL-C}} \\

    \textbf{Models} & \textbf{mAP} & \textbf{mAP@0.5} & \textbf{mAP$^S$}  & \textbf{mAP$^M$} & \textbf{mAP$^L$} & \textbf{mAP} & \textbf{mAP@0.5} & \textbf{mAP$^S$}  & \textbf{mAP$^M$} & \textbf{mAP$^L$} & \textbf{mAP} & \textbf{mAP@0.5} & \textbf{mAP$^S$}  & \textbf{mAP$^M$} & \textbf{mAP$^L$} & \textbf{mAP} & \textbf{mAP@0.5} & \textbf{mAP$^S$}  & \textbf{mAP$^M$} & \textbf{mAP$^L$}\\

    \shline \multicolumn{1}{c}{} & \multicolumn{5}{c|}{\vspace*{-8pt}} & \multicolumn{5}{c|}{} & \multicolumn{5}{c|}{} & \multicolumn{5}{c}{} \\

    LLaVA-1.5-7B 
    & \phantom{2}0.3 & \phantom{2}1.0 & 0.0 & \phantom{2}0.0 & \phantom{2}0.6       %Sketchy
    & \phantom{2}0.3 & \phantom{2}1.0 & 0.0 & \phantom{2}0.0 & \phantom{2}0.6       %QuickDraw
    & \phantom{2}0.5 & \phantom{2}1.8 & 0.0 & \phantom{2}0.0 & \phantom{2}1.0       %TuBerlin
    & \phantom{2}0.4 & \phantom{2}1.4 & 0.0	& \phantom{2}0.1 & \phantom{2}0.7       %SketchVCL-C

    \\
    OneVision
    & \phantom{2}0.3 & \phantom{2}1.2 & 0.0 & \phantom{2}0.1 & \phantom{2}0.5      %Sketchy
    & \phantom{2}0.6 & \phantom{2}1.8 & 0.0 & \phantom{2}0.0 & \phantom{2}1.0      %QuickDraw
    & \phantom{2}0.3 & \phantom{2}1.5 & 0.0 & \phantom{2}0.1 & \phantom{2}0.7       %TuBerlin
    & \phantom{2}0.6 & \phantom{2}1.6 & 0.0 & \phantom{2}0.0 & \phantom{2}1.1       %SketchVCL-C
    \\
    
    DeepSeek-VL2-small
    & \phantom{2}0.4 & \phantom{2}1.2 & 0.0 & \phantom{2}0.1 & \phantom{2}0.7       %Sketchy
    & \phantom{2}0.5 &            1.6 & 0.0 & \phantom{2}0.0 & \phantom{2}0.8       %QuickDraw
    & \phantom{2}0.6 &            2.2 & 0.0 & \phantom{2}0.0 & \phantom{2}1.2       %TuBerlin
    & \phantom{2}0.5 &            1.4 & 0.0 & \phantom{2}0.0 & \phantom{2}1.0       %SketchVCL-C
    \\

    Molmo-7B-D 
    & \phantom{2}0.2 & \phantom{2}0.8 & 0.0 & \phantom{2}0.1 & \phantom{2}0.4       %Sketchy
    & \phantom{2}0.4 & \phantom{2}1.3 & 0.0	& \phantom{2}0.1 & \phantom{2}0.9       %QuickDraw
    & \phantom{2}0.2 & \phantom{2}1.0 & 0.0	& \phantom{2}0.1 & \phantom{2}0.4       %TuBerlin
    & \phantom{2}0.2 & \phantom{2}0.7 & 0.0	& \phantom{2}0.0 & \phantom{2}0.4       %SketchVCL-C
    \\

    \cellcolor{tablegray}\textcolor{molmocolor}{O3SLM-7B (Ours)} 
    & \cellcolor{tablegray}9.9 & \cellcolor{tablegray}21.3& \cellcolor{tablegray}0.6 & \cellcolor{tablegray}6.1 & \cellcolor{tablegray}23.6
    & \cellcolor{tablegray} 6.7 & \cellcolor{tablegray} 13.7 & \cellcolor{tablegray} 0.3 & \cellcolor{tablegray} 4.4 & \cellcolor{tablegray} 18.5
    & \cellcolor{tablegray}8.4 & \cellcolor{tablegray}17.6 & \cellcolor{tablegray}0.4 & \cellcolor{tablegray}4.9 & \cellcolor{tablegray}21.6
    & \cellcolor{tablegray} 5.1 & \cellcolor{tablegray} 11.1 & \cellcolor{tablegray} 0.3 & \cellcolor{tablegray}\phantom{2}3.7 & \cellcolor{tablegray} 15.8
    \\
    \midrule
    
    LLaVA-1.5-13B
    & \phantom{2}0.6 & \phantom{2}2.2 & 0.0 & \phantom{2}0.0 & \phantom{2}1.0       %Sketchy
    & \phantom{2}0.3 & \phantom{2}1.0 & 0.0 & \phantom{2}0.0 & \phantom{2}0.5       %QuickDraw
    & \phantom{2}0.5 & \phantom{2}1.7 & 0.0 & \phantom{2}0.0 & \phantom{2}0.9       %TuBerlin
    & \phantom{2}0.4 & \phantom{2}1.2 & 0.0 & \phantom{2}0.0 & \phantom{2}0.6       %SketchVCL-C       
    
    \\

    DeepSeek-VL2 
    & \phantom{2}0.2 & \phantom{2}0.8 & 0.0 & \phantom{2}0.0 & \phantom{2}0.3       %Sketchy
    & \phantom{2}0.2 & \phantom{2}0.8 & 0.0 & \phantom{2}0.0 & \phantom{2}0.5       %QuickDraw
    & \phantom{2}0.3 & \phantom{2}0.9 & 0.0 & \phantom{2}0.0 & \phantom{2}0.6       %TuBerlin
    & \phantom{2}0.1 & \phantom{2}0.4 & \phantom{2}0.0 & \phantom{2}0.0	& \phantom{2}0.3       %SketchVCL-C
    
    \\

    \cellcolor{tablegray}\textcolor{molmocolor}{O3SLM-13B (Ours)}
    & \cellcolor{tablegray}11.9 & \cellcolor{tablegray}23.7 & \cellcolor{tablegray}1.2 & \cellcolor{tablegray}8.6 & \cellcolor{tablegray}25.4
    & \cellcolor{tablegray} 8.5 & \cellcolor{tablegray} 16.8 & \cellcolor{tablegray} 0.5 & \cellcolor{tablegray} 6.1 & \cellcolor{tablegray} 21.3
    & \cellcolor{tablegray}9.5 & \cellcolor{tablegray}19.2 & \cellcolor{tablegray}0.5 & \cellcolor{tablegray}6.4 & \cellcolor{tablegray}23.7
    & \cellcolor{tablegray}6.6 & \cellcolor{tablegray}13.4 & \cellcolor{tablegray}0.2 & \cellcolor{tablegray}5.3 & \cellcolor{tablegray}18.3
    \\
    
    % \bottomrule

    \end{tabular}
      }%
      \caption{\textbf{Performance of Detection in mAP} }
    % \caption{\textbf{Sketch-Based Object Detection.} To evaluate the sketch-based object detection on images COCO val2017, and sketches from four different datasets, specifically: Sketchy, QuickDraw!, TU-Berlin and SketchVCL-C. Following \cite{kuckreja2024geochat}, we report the Acc metric, since it distinguishes across LVLMs better. For completeness we include mAP scores of our model in the supplementary. $\textcolor{molmocolor}{\dagger}$ indicates sketch datasets which are unseen by our model during training, they assess our model's ability to generalize to sketch styles.}
    \label{tab:detect_map}
\end{table*}